\RequirePackage{fix-cm}
\documentclass[twocolumn,natbib]{svjour3}          
\smartqed  
\usepackage{graphicx}
\usepackage{amsmath}
\usepackage{amsfonts}
\usepackage{booktabs}       
\usepackage{xcolor}
\usepackage{algorithm2e}
\RestyleAlgo{ruled}
\usepackage{multirow} 
\usepackage{float} 

\def\ie{{\em i.e.}}
\def\eg{{\em e.g.}}

\newcommand{\figref}[1]{Fig. \ref{#1}}
\newcommand{\tabref}[1]{Tab. \ref{#1}}

\newcommand{\secref}[1]{Section \ref{#1}}
\newcommand{\algref}[1]{Alg. \ref{#1}}

\newcommand{\mc}[1]{\mathcal{#1}}

\newcommand{\bs}[1]{\boldsymbol{\texttt{#1}}}

\usepackage{balance}

\usepackage[colorlinks, citecolor=blue]{hyperref}

\journalname{International Journal of Computer Vision}
\begin{document}

\title{Semantic Contrastive Bootstrapping for Single-positive Multi-label Recognition
}


\author{Cheng Chen$^{\dag}$        \and Yifan Zhao$^{\dag}$ \and Jia Li$^{*}$
}


\institute{
$^{\dag}$: Equal contribution
$^{*}$: Corresponding author
\\
Cheng Chen \at
              State Key Laboratory of Virtual Reality Technology and Systems,
School of Computer Science and Engineering, Beihang
University, China. \\
              \email{chencheng1@buaa.edu.cn}           
           \and
Yifan Zhao \at
              School of Computer Science, Peking
University, China. \\
              \email{zhaoyf@pku.edu.cn} 
           \and
Jia Li \at
              State Key Laboratory of Virtual Reality Technology and Systems,
School of Computer Science and Engineering, Beihang
University, China. \\
              \email{jiali@buaa.edu.cn} 
}

\date{Received: date / Accepted: date}

\maketitle

\begin{abstract}
Learning multi-label image recognition with incomplete annotation is gaining popularity due to its superior performance and significant labor savings when compared to training with fully labeled datasets. Existing literature mainly focuses on label completion and co-occurrence learning while facing difficulties with the most common single-positive label manner. To tackle this problem, we present a semantic contrastive bootstrapping (Scob) approach to gradually recover the cross-object relationships by introducing class activation as semantic guidance. With this learning guidance, we then propose a recurrent semantic masked transformer to extract iconic object-level representations and delve into the contrastive learning problems on multi-label classification tasks. We further propose a bootstrapping framework in an Expectation-Maximization fashion that iteratively optimizes the network parameters and refines semantic guidance to alleviate possible disturbance caused by wrong semantic guidance. Extensive experimental results demonstrate that the proposed joint learning framework surpasses the state-of-the-art models by a large margin on four public multi-label image recognition benchmarks. Codes can be found at \url{https://github.com/iCVTEAM/Scob}.
\end{abstract}

\section{Introduction}\label{sec1}

Recognizing multiple visual objects within one image is a natural and fundamental problem in computer vision, as it provides prerequisites for many downstream applications, including segmentation~\citep{NEURIPS2021_55a7cf9c},
scene understanding~\citep{NEURIPS2018_432aca3a}, and attribute recognition~\citep{Jia_2021_ICCV}. With the help of sufficient training annotations, existing research efforts~\citep{NEURIPS2021_07e87c2f, Chen_2019_CVPR, NEURIPS2020_05f971b5, Zhao_2021_ICCV, Chen_2019_ICCV, 7780620, Huynh_2020_CVPR, 5995734, detr} have undoubtedly made progress via supervised deep learning models. However, annotating all occurrences of candidate objects, especially small ones, is extremely tedious and labor-consuming, which also usually introduces incorrect noisy labels. Recent approaches towards this challenge prefer to use partial weak labels rather than full annotations, making data collecting considerably easier. In addition, \cite{Durand_2019_CVPR} have demonstrated that training sufficient weak labels shows more promising results than those trained with fully labeled but noisy datasets.

Motivated by this huge potential in multi-label learning, representative works tend to learn the co-occurrence correlations between instances~\citep{wu2018multi,https://doi.org/10.48550/arxiv.2112.10941,9207855}. The other line of work attempts to refine the labeling matrix by pretraining on an accurate fully labeled dataset~\citep{JIANG_2018_ICML,Chen_2019_ICCV} or annotating additional negative training samples~\citep{Durand_2019_CVPR}. Nevertheless, these works inevitably fail to handle extreme circumstances when there are extremely few objects labeled in the same image. As the pioneering work in this field, \cite{Cole_2021_CVPR} established the \textit{single positive} setting for multi-label visual recognition, where only one positive label is annotated in each image. As a less-explored task for recognition, \textit{single positive} multi-label learning is a real problem because most existing datasets,~\eg, ImageNet~\citep{deng2009imagenet}, are only labeled with one single label but with multiple objects occurred~\citep{Dimitris_2020_ICML, Wu_2019}.

\begin{figure}[tbp]%
\centering
\includegraphics[width=0.99\linewidth]{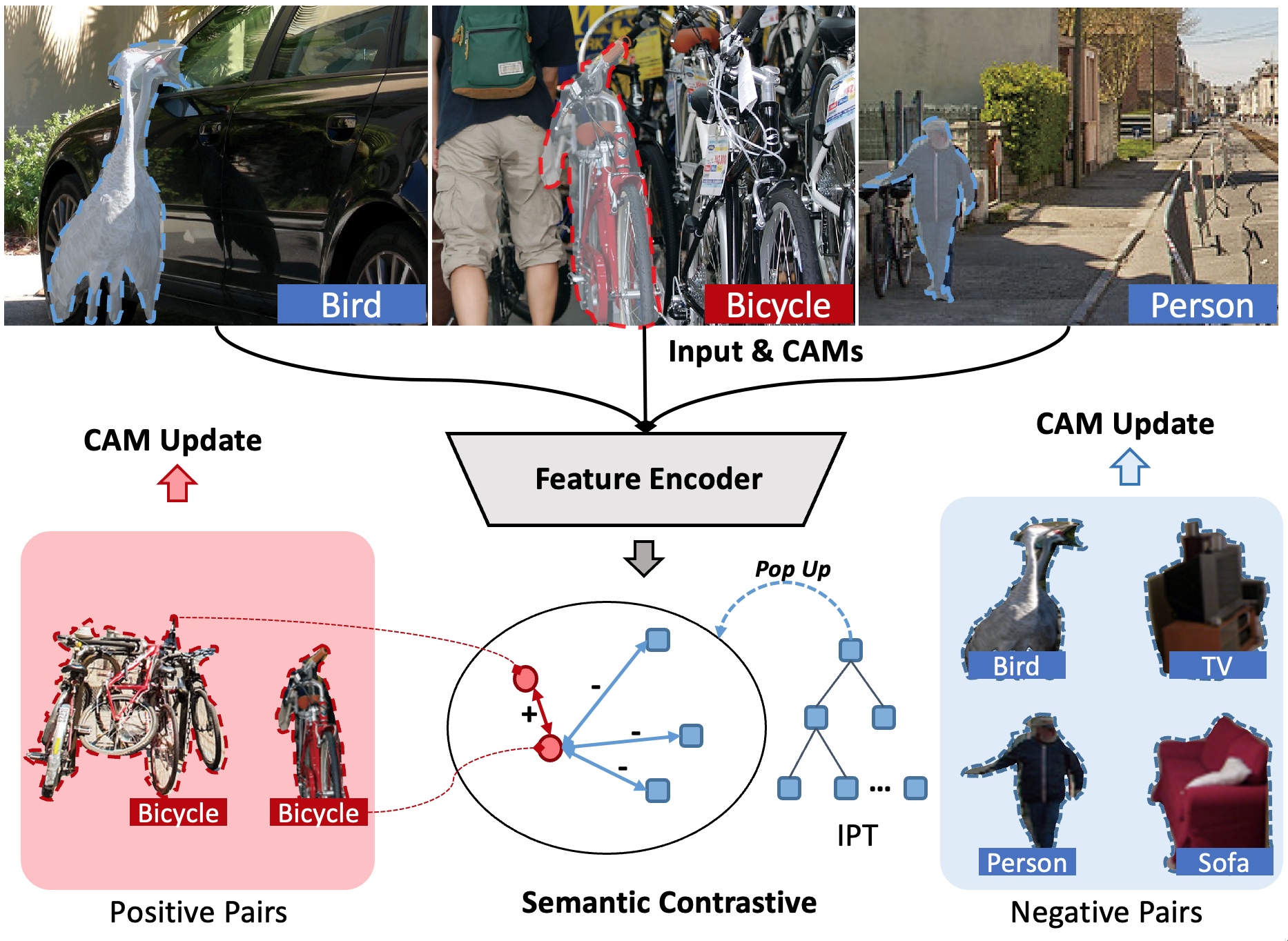}
\caption{  \textbf{Illustrations of our motivation.} Images usually consist of multiple semantic objects while only one is labeled. To mitigate the lack of supervision, we propose a semantic contrastive bootstrapping method that first introduces the gradient-based Class Activation Maps (CAMs) to guide the object-level feature extraction and then build semantic contrastive learning among the positive samples and negative Instance Priority Trees (IPT). To ensure the quality of CAMs, we conduct an EM-based bootstrapping optimization to iteratively update the network features (including IPT) and CAMs.}
\label{fig:motivation}
\end{figure}

Benefiting from the strong fitting ability of deep learning systems, optimizing models with only one positive label would lead to severe negligence on indistinctive objects, while only focusing on the predominant ones. To solve this dilemma, we present a Semantic COntrastive Bootstrapping (Scob) approach, which argues to gradually recover the cross-object relationships from single-positive labels and is constructed from three aspects. \textit{i}) Recent advances in Contrastive Learning (CL) approaches~\citep{NEURIPS2020_d89a66c7} show that deep models have the ability to learn generalized representations without the supervision of manual labels. However, these models are invariably constrained by the discovery of image-level consistencies and discrepancies~\citep{https://doi.org/10.48550/arxiv.2203.06965}, indicating their heavy dependencies on object-centric salient images. For multi-label learning, introducing contrastive learning intuitively would force models to learn ``fake'' relationships between different objects. As in~\figref{fig:motivation}, images only labeled with \textit{bicycle} usually fail to compare with other images owing that the \textit{
person} plays a predominant role in feature extraction. Hence to model the object-level relationships, we introduce the gradient-based Class Activation Maps (CAM)~\citep{gradcamplus} to grasp the object foreground by back-propagating the corresponding class labels, and we then encode them with a Recurrent Semantic Masked Transformer to extract spatial-aware object features in the right side of~\figref{fig:motivation}.

Although the proposed module shows promising benefits for feature extraction, the class activation maps trained by weak labels are usually ambiguous. Regularizing with contrastive learning would lead to accumulative errors when optimizing with such incorrect CAM initialization.  On the other hand, accurate CAM guidance is also highly relied on the network training gradients, which leads to an optimization dilemma between network parameters and semantic CAM guidance.  Toward this end,  we \textit{ii}) develop an instance priority tree to maintain a heap structure for each class, which selects the most confident objects for constructing negative samples. \textit{iii}) We then propose a semantic contrastive bootstrap learning framework to iteratively update the network parameters, and CAM guidance by a generalized Expectation-Maximization model.  With the proposed joint learning framework, experimental evidence demonstrates that our proposed method surpasses the state-of-the-art methods by a large margin in both single-positive and conventional partial label settings.
To sum up, this paper makes the following contributions:

\begin{enumerate}
    \item We introduce Semantic contrastive bootstrapping (Scob) to explore contrastive learning and transform representations in single-positive multi-label recognition, achieving the leading performance on public benchmarks.  
    \item We propose the recurrent semantic masked transformers to purify the object-level information and the instance priority tree for selecting representative negative samples. 
    \item We construct a bootstrap learning framework to formulate network learning and activation updating with the Expectation-Maximization model, revealing its stable convergence in weak label learning systems.
\end{enumerate}

The remainder of this paper is organized as follows: ~\secref{sec:relatedwork} provides the literature review and ~\secref{sec:method} describes
the proposed semantic contrastive bootstrapping approach. Qualitative and quantitative experimental results are reported in~\secref{sec:exp}. ~\secref{sec:conclusion} finally concludes the paper.

\section{Related Works}\label{sec:relatedwork}
In this section, we first present a literature review of multi-label recognition with incomplete labels and then introduce two related techniques which are closely related to our methods,~\ie, the contrastive learning and vision transformers.

\subsection{Multi-Label Recognition with Incomplete Labels}
Multi-label recognition is one of the most fundamental problems in computer vision society~\citep{Tsoumakas_2009_MLOverview} and has attracted increasing research attention~\citep{Chen_2019_CVPR, Zhao_2021_ICCV, Chen_2019_ICCV, 7780620, Huynh_2020_CVPR, 5995734, detr, wu2018multi, Yun_2021_CVPR, Guo_2021_CVPR} in recent years. However, accurate multi-label data are extremely difficult to obtain when given large sets of candidate labels or images. Recent settings on incomplete annotations have attracted much attention. In semi-supervised learning settings, several works~\citep{NEURIPS2021_995693c1, NEURIPS2021_7c93ebe8} assume a subset of the training data is fully labeled while the rest is completely unlabeled. In some partial-label settings, each image is associated with a candidate set containing a correct positive label and many negatives~\citep{NEURIPS2021_217c0e01, https://doi.org/10.48550/arxiv.2201.08984}, whereas in others only a small percentage of labels is known for each image. To solve this problem, several works proposed to handle difficulties meeting in recognition with incomplete labels. \cite{Cole_2021_CVPR} propose to restore labels of the training data as distribution regularization. \cite{NEURIPS2021_10c272d0} propose to learn the correlations among multi-label instances. \cite{JIANG_2018_ICML} propose to clean the noise introduced by missing labels. The other line of works employ a matrix completion algorithm \citep{NIPS2011_65a99bb7, NIPS2013_e58cc5ca,Chen_2019_ICCV} to fill in the missing labels or learn the semantic features between instances and recover the missing labels by the similarity~\citep{Chen_2019_ICCV, https://doi.org/10.48550/arxiv.2112.10941, 10.1007/978-3-319-46448-0_50,Pu2022SARB}. Some weakly-supervised works \citep{10.1145/3240508.3240567, Ge_2018_CVPR, Gao2021LearningTD, Song2021, Zhang2019} also focus on extracting object-level features to enhance the recognition or detection on labeled datasets. However, most of these methods assume that at least a certain percentage of labels is known for each training image, or additional information is provided for learning, which is usually infeasible for the commonly-used single-positive dataset.

In our work, we explore the \textit{single positive} multi-label learning proposed by~\cite{Cole_2021_CVPR}, where only a single positive label is provided for each training image. The single positive label dataset can only provide very little information for multi-label classification training. To overcome it, we leverage contrastive learning to provide more information from instance disambiguation. 
This \textit{single positive} multi-label has significant advantages to collect a large single-label dataset, which can lead to better performances \citep{Durand_2019_CVPR}. It is also easier for human annotators to mark the presence of only a class than notice multiple different presenting classes or what is absent from various images \citep{Wolfe2005}.

\subsection{Contrastive Learning}
Contrastive learning~\citep{NEURIPS2020_d89a66c7} is widely used in self-supervised visual representation learning, which aims at learning representations by distinguishing different images. \cite{He_2020_CVPR} propose a momentum encoder and improve the negative samplings with a novel queue structure. \cite{byol} propose the BYOL model further proves that it is possible to apply contrastive learning with only positive samples. Meantime, \cite{chen2020simple} show that a large enough batch in the training phase is equivalent to the memory bank.
\cite{https://doi.org/10.48550/arxiv.2201.08984} introduce contrastive learning to partial label learning \citep{NEURIPS2021_217c0e01}.
It migrates contrastive learning to mitigate label disambiguation, which is a core challenge of the task. \cite{Liu2019} leverage contrastive learning to improve the distribution of hash of different images in Hamming space. Nevertheless, these methods heavily rely on the training data to satisfy semantic consistency~\citep{https://doi.org/10.48550/arxiv.2203.06965} or sufficient object-central images for representation learning, which are usually infeasible in multi-label datasets. On the other hand, conventional contrastive learning methods tend to construct the generalized representation in an unsupervised manner. This learning mechanism faces a huge dilemma in discovering objects with specific semantic meanings and severely neglects the semantic information of multi-object scenarios.

Recent works also explore the unsupervised object mask with unsupervised learning, For example, DETReg~\citep{bar2022detreg} relies on the pre-trained agnostic object detectors DETR as coarse supervision, and FreeSOLO~\citep{wang2022freesolo} aims to use the predictions of pre-trained network parameters as coarse segmentation masks. Both of these methods do make contributions to focus on the main objects (\ie, relying on object-centric data), but neglect the contrastive relationship between different semantic categories, which still face great challenges in the multi-label classification tasks.

\subsection{Vision Transformers}
Different from CNNs with inherently limited receptive fields, Transformer~\citep{https://doi.org/10.48550/arxiv.1810.04805} is a new structure widely used in natural language processing tasks, which captures the global intrinsic features and relations with self-attention mechanism~\citep{NIPS2017_3f5ee243}. Recent research has indicated that Transformer architectures show great potential in promoting computer vision applications. For example, \cite{NEURIPS2021_64517d84, dosovitskiy2021an} split 2D images into a number of patches and use transformer to produce image features. \cite{NEURIPS2021_3bbfdde8, https://doi.org/10.48550/arxiv.2103.14030, NEURIPS2021_4e0928de} build new transformer suitable for vision tasks. \cite{NEURIPS2021_6dbbe6ab} propose a method based on transformer for self-supervised visual representation learning. \cite{Zhao_2021_ICCV} use transformer to capture long-term contextual information. Transformer has also shown its successes in solving cross-modal tasks in computer vision \citep{Shin2022}. In our work, we resort to our proposed semantic mask transformer to discover object-level features and maintain the semantic consistency on multi-object images.

\begin{figure*}[t]
    \centering
    \includegraphics[width=0.9\linewidth]{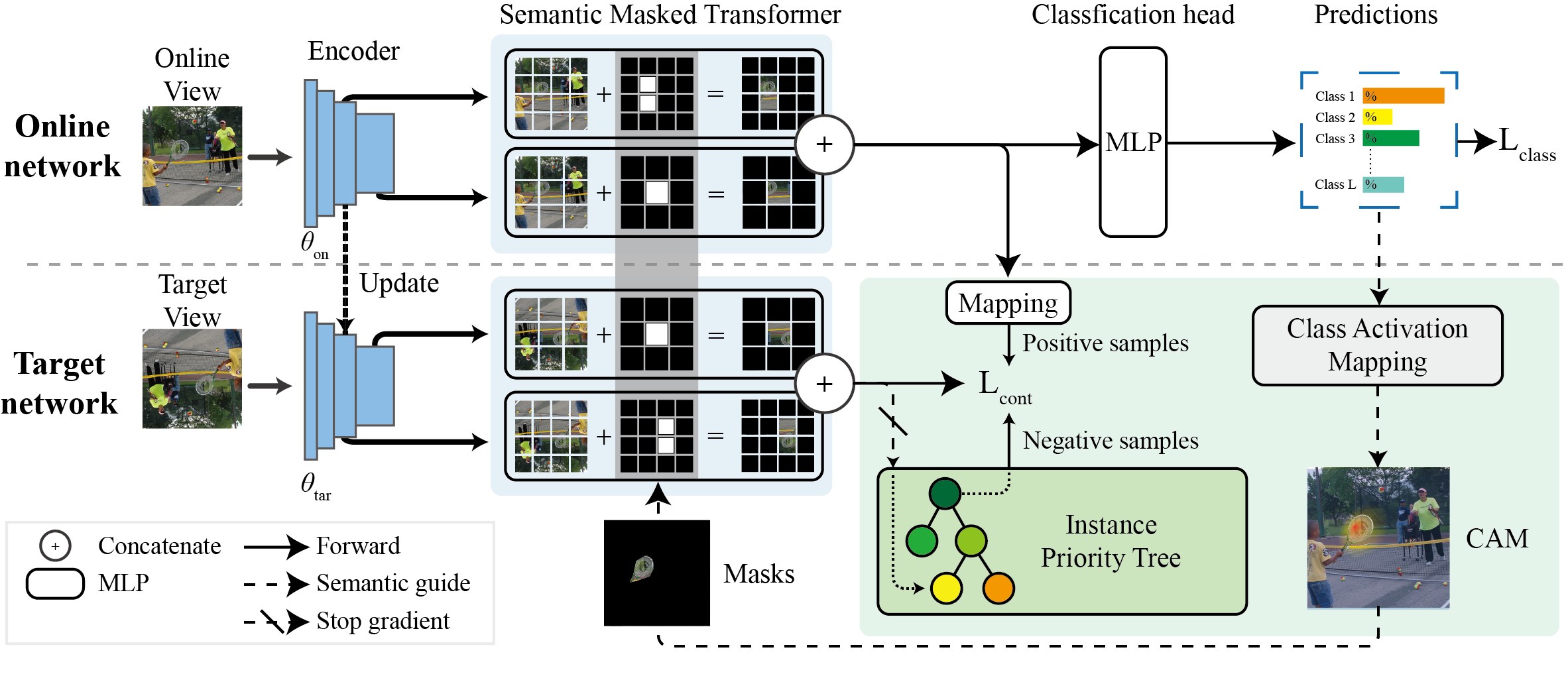}
    \caption{\textbf{The pipeline of the proposed Semantic contrastive bootstrapping (Scob).} Our framework first extracts the object-level class activation maps as feature extraction guidance for Semantic Masksed Transformer (SMT) and then builds a contrastive semantic relationship among the positive samples and negative samples from Instance Priority Tree (IPT). With the previously optimized networks, the object-level CAMs are then calculated by the gradient flow of ground truth labels. The overall bootstrapping framework iteratively optimizes the network parameters and object CAMs. }
    \label{fig:netoverview}
\end{figure*}

\section{Approach}\label{sec:method}
\subsection{Problem Formulation}

\textbf{Single-positive multi-label learning} Let $\mathbf x$ be the inputs sampled from image space $\mathcal X$ and
$\mathbf y \in \{0,1\}^L$ be the associated full labels of input $\mathbf x$, where $L$ is the length of predefined label coding space $\mc{Y}$. In the single-positive learning, the provided annotation $\mathbf{z}$ is randomly sampled from the positive labeling space $\mc{Y}^{+}$. Each image is only annotated with one positive label $\mathbf{z}_n=1$, indicating the objects of the $n$th category exist in image $\mathbf{x}$. While $\mathbf{z}_j=0$ for $j=\{1,2,3,\dots,L\} \backslash \{ n \}$ indicates that the other labels are unknown and could be positive or negative either. For other partial label learning settings, $\mathbf{z} \in \mc{Z}$ can provide more than one label for easier learning. To sum up, our objective is to find a mapping function $\mc{R}: \mc{X} \mapsto \mc{Z}$ with partially annotated data $\mathbf{z}\in \mc{Z}$ for predicting in fully labeling space $\mc{Y}$:
\begin{equation}
 \arg\min_{\Theta} \mathbb{E}_{(\mathbf{x},\mathbf{y})\in \mc{X}\times\mc{Y}}\xi(\mc{R}(\mathbf{x};\Theta),\mathbf{y}),
\end{equation}
where $\Theta$ is the model parameters for mapping function and $\xi: \{0,1\}^L \times \mc{Y}$ is the evaluation criterion~\eg, mAP.

\textbf{Intuitions and framework overview}
As aforementioned, one crucial problem caused by single-positive labeling is that most objects are not iconic or salient in images, hence directly constructing image-level relationships would introduce noisy features caused by other objects. Intuitively, we propose to build an object-level relationship instead of the image levels, which relies on two meaningful modules: 1) recurrent semantic masked transformers to extract purified object-level features; 2) contrastive representation with instance priority trees to select representative negative samples.

\subsection{Recurrent Semantic Masked Transformer}
As the prerequisites for constructing object-level relationships, the major motif of our model is to discover the localization of objects. As there are usually multiple objects occurring in one scenario, it would lead to a severe inductive bias for deep models confusing contextual objects related to the labeled target. For example, models would take the \textit{bicycle} as a part of \textit{person} objects as they are usually present simultaneously in one image. To amend this learning bias during training, we present recurrent semantic masked transformers with Class Activation Maps (CAM) to decompose clusters of multiple objects as separate identities.

As proposed in the field of natural language processing, Transformer models~\citep{https://doi.org/10.48550/arxiv.1810.04805} have the intrinsic characteristic to model the relationships of contextual information. We first split high-level features of ResNet backbones as $W\times H$ patches and then feed them into the Transformer network for modeling contextual relationships and constraining the features to concentrate on regions corresponding to object classes. Beyond this self-attention modeling manner, we then introduce the gradient-based class activations~\citep{gradcamplus} as semantic masks, which are back-propagated by gradients from the last two stages of ResNet backbones. Assume the backbone feature $\mathbf{F} = \Phi(\mathbf{x};\theta_b) \in \mathbb R^{H W\times K} $ is extracted from ResNet stages with parameter $\theta_b \in \Theta$. Hence each gradient feature map corresponded to class $c$, $\mathbf G^{c} \in \mathbb R^{H \times W}$, can be back-propagated by predicted confidence score $\mathbf{p}_{c}$. 
The gradient-based activation $\mathbf G^{c}$ can be formally presented as:
\begin{equation}\label{eq:gradient}
\mathbf{G}^c= \bs{ReLU}(\frac{1}{K}\sum_{k=1}^{K} \underbrace{\frac{1}{WH}\sum_{i=1}^{WH}\frac{\partial \mathbf{p}_{c}}{\partial \mathbf{F}^{k}_{i}}}_{\text{class-related weights}} \cdot \underbrace{\vphantom{\frac{1}{WH}\sum_{i=1}^{WH}\frac{\partial \mathbf{p}_{c}}{\partial \mathbf{F}^{k}_{i}}}\mathbf{F}^{k}}_{\text{feature map}} ).
\end{equation}
To avoid noisy features during this process, we only select the single-positive label to calculate class activation maps. For each location $(i,j)$, we then calculate CAMs $\mathbf M^{c}$ related to class $c$:
\begin{align}\label{eq:cam}
    \mathbf M^{c}_{i,j}&= \bs{ind}\left[ \left( \frac{1}{l^2} \sum_{u=i-\lfloor \frac l2 \rfloor}^{i+\lfloor \frac l2 \rfloor}\sum_{v=j-\lfloor \frac l2 \rfloor}^{j+\lfloor \frac l2 \rfloor}\mathbf{G}^c_{u,v} \right) \geq \gamma_{\mathrm{cam}} \right],
\end{align}
where $l=\frac Hs$ is sampled with window size $s$, $\gamma_{\mathrm{cam}}$ is the threshold hyperparameter, and $\bs{ind}[\cdot]$ is the indicator function, which is $1$ when condition is true, otherwise $0$. One dilemma in solving Eqns.~\eqref{eq:cam} and~\eqref{eq:gradient} is that these activations can only be calculated after back-propagation, while at this moment the network is not forward-propagated to obtain the confidence scores $\mathbf{p}_{c}$. Hence to solve this dilemma, we introduce a recurrent scheme by using the activations of the $(t-1)$th iteration as guidance for the next forward propagation. On the other hand, inspired by the masked learning trend~\citep{he2021masked}, we employ the masked multi-head attention mechanism to obtain object features of the $t$th iteration $\mathbf{H}_{i,j}^{c}(t)$, which are highly responded to class probabilities:
\begin{align}\label{eq:transformer}
\begin{split}
    \mathbf{H}_{i,j}^{c}(t)&=\bs{Attention}( \mathbf{W}_{qry}\mc{M}(\mathbf{F}_{i,j}(t)) \\
    &\quad\quad ,\mathbf{W}_{key}^{\top}\mc{M}(\mathbf{E}_{i,j}(t))^{\top} ) \cdot \mathbf{W}_{val}\mathbf{F}_{i,j}(t),
\end{split}
\\
\begin{split}
    \mc{M}(\mathbf{F}_{i,j}(t)) &= (\mathbf{F}_{i,j}(t)+\mathbf{\mathbf{\Delta}}(i,j)) \\
    &\quad\quad \cdot \mathbf (1-\mathbf{M}^{c}_{i,j}(t-1)),
\end{split}
\end{align}
where $\mathbf{W}_{\{\cdot\}}$ is the learnable attention weights and $\mathbf{\Delta}(\cdot): \mathbb{N}^{W\times H} \rightarrow \mathbb{R}^{1} $ is the learnable positional encoding.
$\mathbf{M}^{c}$ has active regions, where the values are $1$, indicating the foreground, therefore we use $(1-\mathbf{M}^{c}_{i,j}(t-1))$ to mask the background information during learning process. In this manner, the extracted features $\mathbf{H}_{i,j}^{c}(t)$ in each multi-head attention show a high response to the specific semantic classes, serving as perquisites for the relationship discovery among different object instances. The detailed network implementation is elaborated in Appendix.

\begin{figure}[tb]
    \centering
    \includegraphics[width=0.8\linewidth]{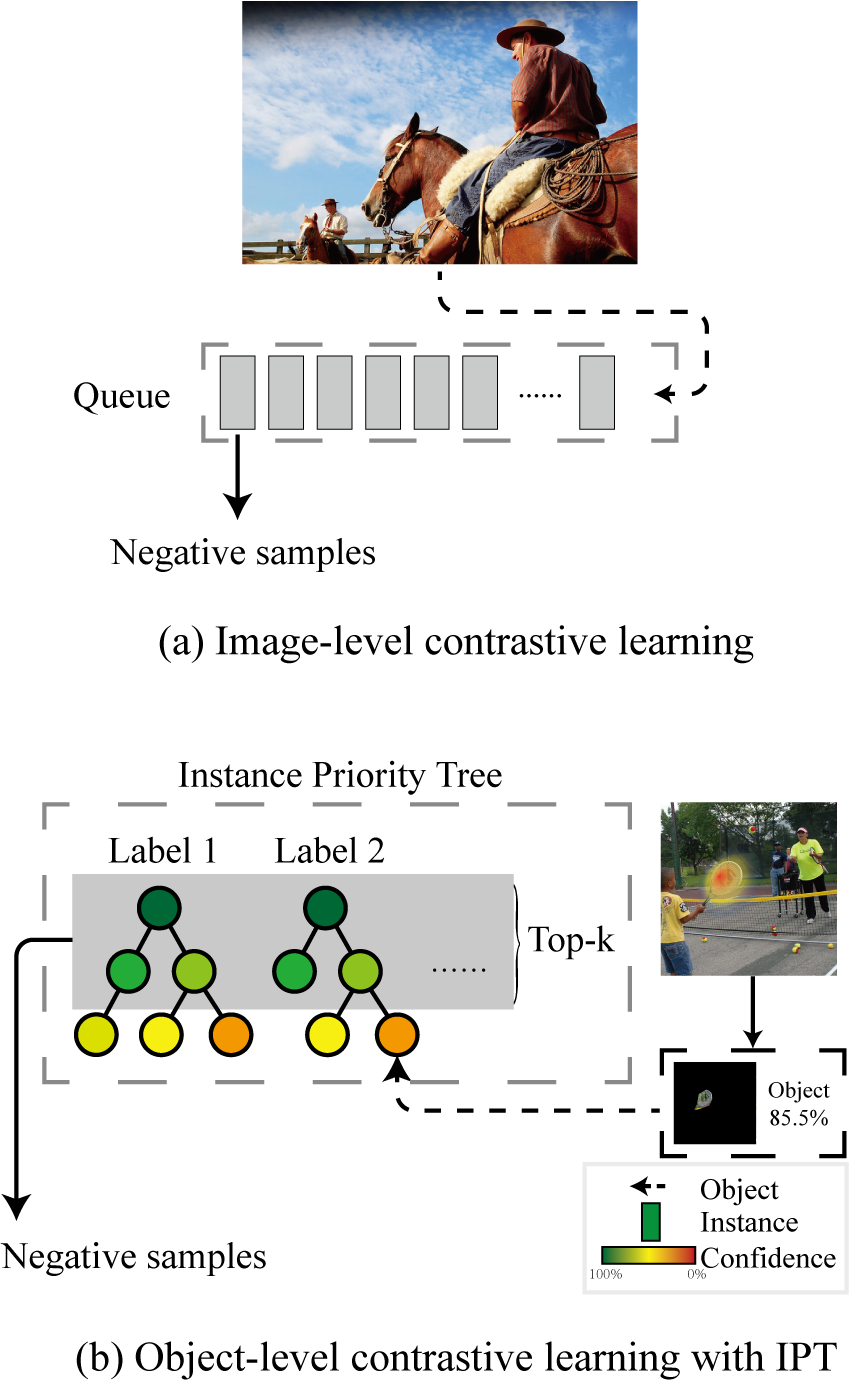}
    \caption{\textbf{The contrastive learning pattern at different levels.} (a) Image-level contrastive learning~\citep{He_2020_CVPR,byol}: introducing the queue structure to collect image-level instances as negative samples.
    (b) Our instance priority tree: the images are masked and instances are inserted into the instance priority tree with confidence. By operating the heaps, the instance priority tree always selects the most confident instances to construct the negative samples, which are used in object-level contrastive learning.}
    \label{fig:ipq}
\end{figure}

\subsection{Instance Priority Tree for Contrastive Learning}
\textbf{Contrastive learning for multi-label recognition} Recent advances~\citep{NEURIPS2020_d89a66c7} in contrastive learning demonstrate that using additional contrastive constraints could help the generalization ability of model learning by alleviating overfitting issues. The main objective in prevailing contrastive learning methods is to construct \textit{positive} and \textit{negative} instances to learn semantic consistency across images. However, their success heavily relies on object-centric images and learning general concepts from a large amount of data, which are usually unavailable for the multi-label dataset. Hence our goal is to construct object-level relationships for contrastive learning with our proposed object-level semantics. Our contrastive architecture follows the state-of-the-art works, including MoCo~\citep{He_2020_CVPR} and BYOL~\citep{byol}, which is composed of an online network $\mc{T}_{on}$ and target network $\mc{T}_{tar}$.

Given an image $\mathbf x_n$ associated with a single-positive label $\mathbf z_n=\{0,1\}^L$, we randomly select the positive samples $\mathbf x_t^{+}$ from $\mc{X}^{+}_{n} =\{\mathbf{x}_i; \mathbf{z}_i = \mathbf{z}_n \}_{i=1}^{N} $, and negative ones $\mathbf x_t^{-}$ from $\mc{X}^{-}_{n}, \mc{X}^{-}_{n}= \mc{X}_n \backslash \mc{X}^{+}_{n} $ and then generate two distinctive samples by view augmentation $\bs{Aug}$. The augmented views are then encoded with the proposed transformers to obtain feature $\widetilde{\mathbf{H}}$: 
\begin{align}\label{eq:augment}
\widetilde{\mathbf{H}}_{on}&=\mc{T}_{on}(\bs{Aug}_o(\mathbf x_n);\theta_{on}), \\
\begin{split}
    \widetilde{\mathbf{H}}_{tar}^{\{+,-\}}&=\mc{T}_{tar}(\bs{Aug}_t(\mathbf {x}_t^{\{+,-\}});\theta_{tar}), \\
    &\qquad\qquad\qquad \mathrm{where}~\mathbf x_t^{+} \in \mc{X}^{+}_{n}, \mathbf x_t^{-} \in \mc{X}^{-}_{n}.
\end{split}
\end{align}
After obtaining contrastive samples, we then attach a classification head $g$ to predict multi-label probability $\mathbf p_n$ for image $\mathbf x_n$.

\textbf{Instance priority tree} Constructing object-level contrastive constraint requests representative object features of each class. However, learning with weak image-level labels usually gets inferior CAMs for object localization, especially for those with co-occurrence relationships. For example, it is hard to disentangle iconic CAMs for \textit{horse} and \textit{person} in a riding scenario (\figref{fig:ipq} a). Hence we propose a heap structure to select representative samples from each class as negative instances for contrastive learning. 
As in~\figref{fig:ipq} b, for each class $c$, we maintain a complete binary tree graph $\mathbb{G}=\{\mc{E}, \mc{N}\}$ for selecting the samples with high confidence. Each node $\mathbf{N}$ is composed of masked semantic node feature $\widetilde{\mathbf{H}}$ with corresponding activation confidence scores $\mathbf{s}$. And each edge $e \in \mc{E}$ indicates an affiliation relationship where the parent node is with higher confidence than the leaf ones.  To avoid noisy features, we only use the object features that are supervised by single-positive ground truth to construct the proposed tree structure.

By leveraging this tree structure, we then obtain the top-$K$ confident negative samples for each class, resulting in an affordable complexity of $\mc{O}(K\log N_c)$. $N_c$ is the node number of the priority tree of the $c$-th class, which is much larger than the batch size. 
With the constructed representative samples, the contrastive InfoNCE loss~\citep{Oord2018RepresentationLW} could be adapted as:
\begin{equation}\label{eq:contrastloss}
\begin{split}
    &\mathcal L_{\mathrm{cont}}(\widetilde{\mathbf H}, \widetilde{\mathbf H}^{\{+,-\}},\mathcal N) \\
    &=-\lambda_c \frac{1}{1+\lvert\mathcal{N} \backslash \mathcal{N}^i\rvert} \\
    &\quad\cdot \log\frac{\exp(\widetilde{\mathbf H} \cdot \widetilde{\mathbf H}^+/\tau)}{\exp(\widetilde{\mathbf H} \cdot \widetilde{\mathbf H}^+/\tau)+\sum_{\widetilde{\mathbf H}^- \in \mathcal N \backslash \mathcal N^i} \exp (\widetilde{\mathbf H} \cdot \widetilde{\mathbf H}^-/\tau)},
\end{split}
\end{equation}
where $\widetilde{\mathbf{H}}^{\{+,-\}}$ come from the target network indirectly, $\mathcal N \backslash \mathcal N^i$ are samples maintained by $\mathbb G$ except those labeled with $i$. $\tau$ and $\lambda_c$ is temperature and weighting parameter. $\widetilde{\mathbf H}$ is then used to update instance priority tree $\mathbb G$. By Eqns.~\eqref{eq:augment} $\sim$ \eqref{eq:contrastloss}, our object-level contrastive learning then regularizes the network to be aware of representative object features for each class.

\section{Bootstrapping Optimization with Expectation-Maximization}
For each optimization tuple $(\mathbf{x},\mathbf{M},\mathbf{z})$ of the input image, CAM guidance, and observed label, one natural concern is that class activation $\mathbf{M}$ can only be obtained by prediction $\mathbf{p}$ via back-propagation processes, which can not be obtained afore the optimization. In other words, as activation $\mathbf{M}$ serves as the semantic guidance for feature extraction, a bad initialization of $\mathbf{M}$ would lead to the failure of whole learning optimization. For such a ``chicken-and-egg'' conundrum, we thus pursue a bootstrapping optimization with Expectation-Maximization (EM) modeling. For simplicity, if we ignore terms not depending on $\Theta$,~\eg, distribution regularization, the expected log-likelihood can be defined as:
\begin{equation}
\begin{split}
    &\mc{Q}(\Theta^{t},\Theta^{t-1}) \\
    &\quad = \sum_{\mathbf{M}}\mc{P}(\mathbf{M}\mid\mathbf{x},\mathbf{z}; \Theta^{t-1})\log \mc{P}(\mathbf{x},\mathbf{M};\Theta^{t}).
\end{split}
\end{equation}
For the hidden activation map $\hat{\mathbf{M}}$ using the $t$-1th model, we then simplify Eqns.~\eqref{eq:gradient} and~\eqref{eq:cam} as function $\bs{CAM}$, thus the \textbf{E-step} optimization on $\hat{\mathbf{M}}$ is:
\begin{align}
\begin{split}
    \hat{\mathbf{M}} &= \arg\max_{\mathbf{M}} \mc{P}(\mathbf{M}\mid\mathbf{x};\Theta^{t-1}) \\
    &\approx \bs{CAM}( \frac{\partial g(\Phi(\mathbf{x};\Theta^{t-1}))}{\partial \Phi(\mathbf{x};\Theta^{t-1})}\Phi(\mathbf{x};\Theta^{t-1})),
\end{split}
\end{align}
where $g(\cdot)$ denotes the classification head. Instead of directly calculating the most confidence region, we approximate $\hat{\mathbf{M}}$ by the Gradient CAM~\citep{gradcamplus} via back-propagating the prediction probability $\mathbf{p}$.

In the \textbf{Maximization-step} of our framework, the $\mc{Q}$ optimization is composed of two parts, ~\ie, optimization on the online network $\mc{Q}(\cdot;\theta_{on})$ and target network $\mc{Q}(\cdot;\theta_{tar})$.
For the main online network, we introduce auxiliary distribution constraints to encourage the discovery of multiple objects. In each mini-batch $B$ of size $\mathrm{b}$, we denote $\mathbf P_B,\mathbf Z_B,\widetilde{\mathbf Y}_B \in \mathbb{R}^{b \times L}$ as predictions and labels of batch $B$, and $\mathbf p_n,\mathbf z_n, \widetilde{\mathbf y}_n$ indicate rows of them respectively.
Following~\cite{Cole_2021_CVPR}, the single-positive multi-label classification loss is composed of three parts: 1) single-positive binary cross-entropy (BCE) $\mathcal L_{\mathrm{BCE}}^+$: only calculated with one-hot positive label $\mathbf{z}$; 2) standard BCE loss $\mathcal L_{\mathrm{BCE}}$: updated using the estimated multi-label $\widetilde{\mathbf y}_n$ stored in the estimator matrices; 3) distribution regularization: maintaining the averaged prediction share similarities with distributions of training data. For batch-wise predictions $\mathbf{P}_B$ and estimated weak label $\widetilde{\mathbf{Y}}_B$ which are predicted by the network and the estimator, it has the following form:
\begin{align}
\begin{split}
    \mathcal L_{\mathrm{sp}}&(\mathbf P_B \mid \widetilde{\mathbf Y}_B;\theta_{on}) \\
    &= \frac{1}{|B|}\left(\sum_{n \in B}   \underbrace{\mathcal L_{\mathrm{BCE}}(\mathbf p_n,\mc{S}(\widetilde{\mathbf y}_n))}_{\text{pseudo multi-label}} + \sum_{n \in B} \underbrace{\mathcal L_{\mathrm{BCE}}^+(\mathbf p_n,\mathbf z_n)}_{\text{ single-positive}} \right) \\
    &\qquad + \underbrace{\left( (\hat k(\mathbf P_B)-k)/{L} \right)^2}_{\text{distribution constraint}},
\end{split}
\end{align}
where $\mc{S}(\cdot)$ denotes the stop-gradient function and $\hat k,k$ are the expectation of number of positive labels per image (refer to \cite{Cole_2021_CVPR} for details). Switching the arguments $\mathbf F_B, \widetilde{\mathbf Y}_B$ and combining them, we get
\begin{equation}
    \begin{split}
        &\mathcal L_{\mathrm{class}}(\mathbf F_B,\widetilde{\mathbf Y}_B;\theta_{on}) \\
        &\qquad =\frac{\mathcal L_{\mathrm{sp}}(\mathbf F_B \mid \widetilde{\mathbf Y}_B ;\theta_{on})+\mathcal L_{\mathrm{sp}}(\widetilde{\mathbf Y}_B \mid \mathbf F_B ;\theta_{on})}{2},
    \end{split}
\end{equation}
where in the initialization phase of $\widetilde{\mathbf{Y}}$, we set the probability after the sigmoid function of known labels close to 1 and initialize the probability after the sigmoid of unknown labels into $u$, where $u \sim [0.5-\xi, 0.5+\xi]$ and $\xi=0.3$ empirically following~\cite{Cole_2021_CVPR}. Besides the classification loss, the proposed contrastive loss $\mathcal L_{\mathrm{cont}}(\cdot; \theta_{on})$ in Eqn.~\eqref{eq:contrastloss} also regularizes the network after updating $\hat{\mathbf{M}}$ in the E-step. Following other contrastive learning schemes~\citep{byol}, the target network $\theta_{tar}$ follows momentum updating trend:
\begin{equation}
    \theta_{tar} \gets \alpha \theta_{tar} + (1-\alpha) \theta_{on},
\end{equation}
where $\theta_{tar}$ and $\theta_{on}$ is the parameters of target network $\mc{T}_{tar}$ and online
network $\mc{T}_{on}$, $\alpha \in [0,1)$ is a momentum coefficient. Our bootstrapping framework is elaborated in~\algref{alg:framework}.

\SetKwComment{Comment}{/* }{ */}
\begin{algorithm}[t]
\caption{Bootstrapping Optimization with Expectation-Maximization}\label{alg:framework}
\KwData{Training data $\mathcal X$ with associated label space $\mathcal Z$, online network $\theta_{on}$, target network $\theta_{tar}$.}

Initialize $\mathbb G^c, c \in \{1,2,\dots,C\}, t=0$\; 
Initialize $\Theta^{0} \gets \bs{optimizer}(\Theta,\nabla_{\Theta}\mc{L}_{\text{class}})$\;
\While{not convergence $\bs{or}$ $t <\bs{MaxIter}$}{
    \For{$(\mathbf x_n, \mathbf z_n) \sim B(\mc{X}\times \mc{Z})$}{
        $t \gets t+1$ \;
        Get prediction $\mathbf{P}_{B}=g(\Phi(\mathbf{x};\Theta^{t-1}))$\;
        Obtaining class activation map $\hat{\mathbf{M}}^{t-1} = \bs{CAM}(\mathbf{x};\Theta^{t-1})$ by Eqn.~\eqref{eq:cam} \Comment*[r]{{\color{blue}\textit{E-Step}}} \
        $\mathbb G^{\mathbf{p}_{k}} \gets \mathbb G^{\mathbf{p}_{k}} \cup \{ \mathbf{H}(\hat{\mathbf{M}}_{\mathbf{p}_{k}}), \mathbf{s}_{\mathbf{p}_{k}} \}$ by Eqn.~\eqref{eq:transformer} \; 
        $\widetilde{\mathbf{H}}_{on}=\mc{T}_{on}(\bs{Aug}_o(\mathbf x_n), \hat{\mathbf{M}}^{t-1};\theta_{on})$\;
        $\theta^{t}_{on} \gets \bs{optimizer}(\theta_{on},\nabla_{\theta_{on}}\mc{L}_{\text{class}}(\mathbf{P},\widetilde{\mathbf Y}^{t-1}_B))$, update $\widetilde{\mathbf Y}_B^{t}$\
        \Comment*[r]{{\color{blue}\textit{M-Step}}}\
        $\widetilde{\mathbf{H}}_{tar}^{+}=\mc{T}_{tar}(\bs{Aug}_t(\mathbf {x}_t^{+}), \hat{\mathbf{M}}^{t-1};\theta_{tar})$\;
        Sample top-k negative samples from instance priority tree $ \widetilde{\mathbf{H}}_{tar}^{-} \sim \mathbb{G}^{c}, c \neq \mathbf{z}_n$ \;
       $\theta^{t}_{on} \gets \bs{optimizer}(\theta_{on},\nabla_{\theta_{on}}\mc{L}_{\text{cont}}(\widetilde{\mathbf H}_{on}, \widetilde{\mathbf H}_{tar}^{\{+,-\}}))$\;
    }
    $\theta_{tar}^{t} \gets \alpha \theta_{tar}^{t-1} + (1-\alpha) \theta_{on}^{t}$  
    \Comment*[r]{{\color{blue}\textit{Target Net Updating}}}
}
\end{algorithm}

\begin{table*}[!t]
  \caption{  mAP Comparisons (\%) with the state-of-the-art methods on the VOC 2007/2012, CUB and COCO datasets. $10\%$ Pos. Neg.: training with 10\% of full labels. 1 Positive: single positive setting. *: values are averaged over three runs. The best results on each dataset are highlighted in bold.}
  \label{tab:maptable}
  \centering

  \begin{tabular}{cll|cccc}
    \toprule
    \textbf{Methods}    & \textbf{Backbone}         & \textbf{Labels Per Image} & \textbf{VOC 2007}         & \textbf{VOC 2012 }            & \textbf{CUB}        & \textbf{Microsoft COCO}           \\
    \midrule
    SSGRL           & ResNet-101   & $10\%$ Pos. Neg. & $77.7$            &         -      &   -   & $62.5$          \\
    GCN-ML          & ResNet-101   & $10\%$ Pos. Neg. & $74.5$            &         -      &  -    & $63.8$          \\
    KGGR            & ResNet-101   & $10\%$ Pos. Neg. & $81.3$             &        -      &  -     & $66.6$          \\
    Curriculum labeling & ResNet-101 & $10\%$ Pos. Neg. & $44.7$            &         -      &   -   & $26.7$          \\
    Partial BCE     & ResNet-101   & $10\%$ Pos. Neg. & $80.7$            &         -      &   -   & $61.6$          \\
    SST             & ResNet-101   & $10\%$ Pos. Neg. & $81.5$             &        -      &    -   & $68.1$          \\
    SARB            & ResNet-101    & $10\%$ Pos. Neg. & $83.5$             &        -     &    -    & $71.2$          \\
    ROLE            & ResNet-50   & $1$ Positive    & -            &    $88.2$          &    $16.8$   & $69.0$          \\
    \midrule
    \textbf{Scob} (Ours)*    &  ResNet-50          & $1$ Positive    &  $\mathbf{88.5}$ & $\mathbf{89.7}$ & $\mathbf{20.4}$ & $\mathbf{74.8}$ \\
    \textbf{Scob} (Ours)*    &  ResNet-50          & $10\%$ Pos. Neg. &  $\mathbf{88.9}$ & $\mathbf{89.8}$ & $\mathbf{21.4}$ & $\mathbf{75.2}$ \\
    \bottomrule
  \end{tabular}
\end{table*}

\section{Experiments}\label{sec:exp} 

\subsection{Experiment Setup}\label{sec:dataset}

\textbf{Datasets}
Following the previous studies \citep{Cole_2021_CVPR, https://doi.org/10.48550/arxiv.2112.10941}, we conduct experiments on four representative benchmarks,~\ie, PASCAL VOC 2007~\citep{pascal-voc-2007}, PASCAL VOC 2012~\citep{pascal-voc-2012}, Microsoft COCO 2014~\citep{10.1007/978-3-319-10602-1_48}, CUB-200-2011~\citep{WahCUB_200_2011}, which are fully labeled and widely-used in multi-label image recognition.
PASCAL VOC~\citep{pascal-voc-2012} contains $20$ categories, where each image in the dataset has $1.4$ labels on average. Microsoft COCO~\citep{10.1007/978-3-319-10602-1_48} is the most widely-used and challenging benchmark in multi-label classification tasks, which contains $80$ categories. Each image in the MS-COCO dataset contains 2.9 labels on average. It explores the object recognition task, altering the task from recognizing only the prominent object to understanding multiple objects in holistic scenarios. CUB-200-2011~\citep{WahCUB_200_2011} is the most widely-used dataset for fine-grained visual classification tasks. It has 312 binary attributes per image describing various details, which is more challenging than the first three datasets.

\textbf{Baselines settings}
We adopt the re-implemented ROLE \citep{Cole_2021_CVPR} as our baselines in the following experiments. i) While different from ROLE \citep{Cole_2021_CVPR} using linear initialization, in our baseline model (\ie, Scob w/o all) all proposed modules are replaced with two convolutional layers as a simple classification head and contrastive learning is disabled in training.
ii) To evaluate the effectiveness of our model, we implement the \textit{large-CNN} network with more training parameters by adding convolutional blocks, which is detailed in~\tabref{tab:ablation}.

\textbf{Evaluation metrics} For fair comparisons, we follow \cite{https://doi.org/10.48550/arxiv.2112.10941, Cole_2021_CVPR, Durand_2019_CVPR} and adopt the mean Average Precision (mAP) as evaluation metrics. 
As there are few works reported their results on the single-positive setting, we compare our methods with the partial label learning methods \citep{Chen_2019_ICCV, Chen_2019_CVPR, 9207855, Durand_2019_CVPR, https://doi.org/10.48550/arxiv.2112.10941, Pu2022SARB} on the $10\%$ partial label settings. 
that Scob is proposed in \textit{single-positive multi-label setting} as aforementioned.  Besides the mAP results, we additionally report average overall precision (OP), overall recall (OR), overall F1-score (OF1), per-class precision (CP), per-class recall (CR), and per-class F1-score of some approaches. For fair comparisons, we report the mean values with three random seeds, while the results of other works are reported by their original paper.

\begin{table*}[!t]
  \centering
    \caption{ Detailed comparisons on different evaluation metrics. The baseline model \citep{Cole_2021_CVPR} is re-implemented by ours, which shows slightly higher performances.}
  \begin{tabular}{c|cl|cccccc}
    \toprule
    \textbf{Benchmark}                              & \textbf{Methods} & \textbf{Labels}           & \textbf{OP}     & \textbf{OR}     & \textbf{OF1}    & \textbf{CP}     & \textbf{CR}     & \textbf{CF1}    \\
    \midrule
    \multirow{3}{*}{PASCAL VOC 2007}     & Baseline~\citep{Cole_2021_CVPR}   & $1$ Positive     &  $91.7$ & $77.4$ & $84.0$ & $90.3$ & $74.5$ & $81.6$ \\
                                        & Scob (Ours)   & $1$ Positive     & $90.0$ & $80.2$ & $84.9$ & $86.7$ & $79.2$ & $82.8$ \\
                                         & Scob (Ours)  & $10\%$ Pos. Neg. & $87.3$ & $84.4$ & $85.8$ & $84.4$ & $82.8$ & $83.6$ \\
    \midrule
    \multirow{3}{*}{PASCAL VOC 2012}     & Baseline~\citep{Cole_2021_CVPR}   & $1$ Positive     & $85.2$ & $83.3$ & $84.2$ & $83.9$ & $81.1$ & $82.5$ \\
                                         & Scob  (Ours) & $1$ Positive     & $88.2$ & $82.1$ & $85.0$ & $84.6$ & $81.7$ & $83.1$ \\
                                         & Scob (Ours)  & $10\%$ Pos. Neg. & $87.1$ & $85.2$ & $86.2$ & $85.3$ & $83.5$ & $84.4$ \\
    \midrule
    \multirow{3}{*}{MS COCO 2014} & Baseline~\citep{Cole_2021_CVPR}   & $1$ Positive     & $70.1$ & $68.3$ & $69.2$ & $68.0$ & $63.1$ & $65.5$ \\
                                         & Scob (Ours)  & $1$ Positive     & $81.1$ & $66.2$ & $72.9$ & $78.7$ & $61.1$ & $68.8$ \\
                                         & Scob  (Ours) & $10\%$ Pos. Neg. & $79.3$       & $68.6$       & $73.6$       & $79.7$       &    $62.8$    & $70.2$       \\
    \bottomrule
  \end{tabular}

  \label{tab:more_metrics}
\end{table*}

\begin{table*}[!t]
\caption{Ablation studies of different components. Rec. SMT: Recurrent semantic masked transformers. CL: Contrastive learning. IPT: instance priority tree for negative sampling. $/$: the proposed module is not used. $+$: the proposed module is enabled. w/o CAM: only remove CAM but retain other parts in SMT.}
\label{tab:ablation}
\centering
\begin{tabular}{ccc|cc}
    \toprule
      \textbf{Rec. SMT} & \textbf{CL}    &   \textbf{Sampling}            &  \textbf{Parameter} & \textbf{mAP on MS-COCO}            \\
    \midrule
     $/$ &       $/$     & $/$       & $36.6~\textrm{M}$  & $69.8$           \\
   Large CNN &       $/$      & $/$       & $45.0~\textrm{M}$  & $70.9$           \\
    CAM & $/$   &  $/$                           & $38.3~\textrm{M}$  & $73.8$          \\
     $/$ & $+$          & IPT                    & $36.6~\textrm{M}$   & $68.9$          \\
     w/o CAM & $+$                  &  Random  & $38.3~\textrm{M}$   & $73.2$          \\
    w/o CAM & $+$                  &  IPT           & $38.3~\textrm{M}$   & $73.9$          \\
     CAM & $+$                  &  Random          & $38.3~\textrm{M}$    & $74.3$          \\
     \midrule
    CAM & $+$                  &  IPT     & $38.3~\textrm{M}$ & $\mathbf{74.8}$          \\
    \bottomrule
  \end{tabular}
\end{table*}

\subsection{Implementation Details}\label{sec.implementation}
\textbf{Data preprocessing} Following the single-positive learning setting \citep{Cole_2021_CVPR}, we firstly drop the existing annotations of datasets and randomly retain a positive label per image. While for partial-label settings, following state-of-the-art works \citep{https://doi.org/10.48550/arxiv.2112.10941}, we randomly reserve $10\%$ labels per image for a fair comparison. These operations are performed only once so that the randomly retained label is fixed for all approaches. To evaluate the performance of these approaches, we keep all the labels of the official validation sets. The results are reported on the whole validation set.

\textbf{Training details}  We adopt ResNet-50~\citep{He_2016_CVPR} pretrained on ImageNet~\citep{deng2009imagenet} as backbone for fair comparison. The backbone can be easily replaced with other networks like ViT \citep{ViT}. Following \cite{byol}, images are resized into $448 \times 448$ with data augmentation.
Similar to \cite{detr}, $P$ is implemented as a self-learning positional encoding function.
We adopt the Adam optimizer and train the model for $30$ epochs. The learning rates for the estimator, SMT, mapping layers, and others in the online network are $0.01$, $4 \times 10^{-4}$, $0.01$, $0.001$ respectively.
The batch size is set as $8$. The dimension of features is set as $512$. The hidden dimension of the transformer is set as $2048$. Each transformer unit consists of $2$ layers and each layer has $8$ attention heads. The threshold $\gamma_\mathrm{cam}$ is set as $0.5$. The scalar $k$ is set as $1.5$, $31.4$, and $3.0$ for VOC 2007/2012, CUB, and COCO, following \cite{Cole_2021_CVPR}. The hyperparameters $\lambda_c$ is set as $0.1$. The temperature $\tau$ of contrastive learning loss is set as $1.0$. The momentum factor $\alpha$ is set as $0.999$ following \cite{byol}. The size of $\mathbb G$ is set as $80$. The backbone is frozen during training.

\subsection{Comparison with State-of-The-Art Approaches}

We respectively compare our approach on PASCAL VOC 2007, VOC 2012, COCO and CUB datasets with 8 state-of-the-art methods, including SSGRL~\citep{Chen_2019_ICCV}, GCN-ML~\citep{Chen_2019_CVPR}, KGGR~\citep{9207855}, Curriculum labeling~\citep{Durand_2019_CVPR}, Partial BCE~\citep{Durand_2019_CVPR}, SST~\citep{https://doi.org/10.48550/arxiv.2112.10941}, SARB~\citep{Pu2022SARB}, and ROLE~\citep{Cole_2021_CVPR}.
As there are few works that reported their results on the single-positive setting, we also compare our methods with these aforementioned partial label learning works. 
However, these methods usually rely on the label-aware co-occurrence, which is corrupted for optimization in this single-positive setting.
The mAP results on four benchmark datasets are exhibited in \tabref{tab:maptable}. It is clear that our approach achieves leader-board performance and significantly outperforms SARB~\citep{Pu2022SARB} on both VOC and COCO by $5.8\%$ and $4.1\%$, ROLE~\citep{Cole_2021_CVPR} by $1.6\%$ (VOC 2012) and $5.8\%$ (COCO). Moreover, our approach achieves much higher performance than these leading works by only leveraging much fewer annotations (1 Positive) than those with $10\%$ partial labels. With the increase of annotation in the last two rows, our proposed method shows a notable improvement, verifying the robustness and effectiveness of different propositions of incomplete data annotations.

Besides the most widely-used mAP, we report average overall precision (OP), overall recall (OR), overall F1-score (OF1), per-class precision (CP), per-class recall (CR), and per-class F1-score of the baseline \citep{Cole_2021_CVPR} and Scob on three datasets in \tabref{tab:more_metrics}. Considering the compared methods in the partial label learning or single-positive learning setting do not report values on these metrics, we re-implement the state-of-the-art model ROLE~\citep{Cole_2021_CVPR}, which performs slightly higher results than the paper reports.

\subsection{Performance Analyses}
\textbf{Effect of recurrent Semantic Masked Transformers (SMT)} To ablate SMT, we replace it with several 2D convolutional layers for setting w/o SMT.  From \tabref{tab:ablation}, we can observe that SMT notably improves the performance by incorporating a self-attention mechanism with reasonable masked semantic guidance compared to the baseline,~\ie, from 69.8 to 73.8. By only removing the CAM guidance in SMT (shown in the sixth row), the overall performance decreases by $0.9\%$. This indicates that semantic guidance plays an important role in learning object-level relationships.

To validate our proposed method is mainly improved by additional parameters. In the second row, we also train a \textit{large-CNN} as aforementioned that has much more parameters than ours without contrastive learning and observe that the effect of additional parameters on the performance is very limited, which further clarifies that the improvement is not just due to the additional parameters, but the cooperation of proposed new modules and object-level contrastive learning.

\textbf{Different variants of contrastive regularization} In \tabref{tab:ablation}, our full model substantially outperforms the third row without contrastive learning, which shows the effectiveness of learning object-level contrastive representation. Moreover, the performance of Scob w/o IPT decreases by $0.5\%$, in which a \textit{random} sampling strategy equivalent to the vanilla memory bank used in MoCo~\citep{He_2020_CVPR} is applied, indicating that Instance Priority Tree (IPT) provides notable improvements with representative negative object samples. Contrastive learning helps the framework learn better representations of different labels by object disambiguation. It has been observed that networks can predict random labels, but the same network can fit informative data faster~\citep{Cole_2021_CVPR}. Better representations let the model fit single positive data during training, recovering better weak labels.

\begin{figure}[!t]
\centering
\includegraphics[width=0.85\linewidth]{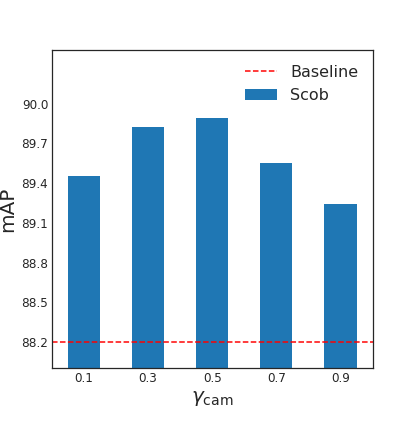} 
\caption{\textbf{mAP performance on VOC12 dataset with different CAM threshold $\gamma_{\mathrm{cam}}$.}}
\label{fig:ablation_gamma_cam}
\end{figure}

\begin{figure}[!t]
    \centering
    \includegraphics[width=0.85\linewidth]{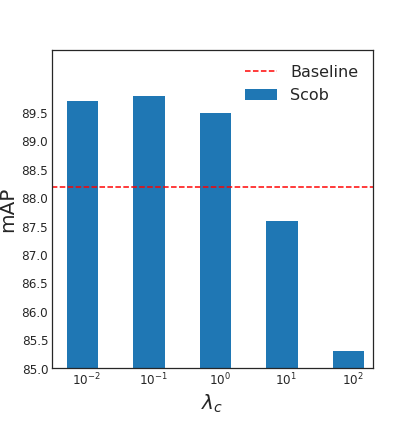}
    \caption{\textbf{mAP performance on VOC12 dataset with different contrastive learning coefficient $\lambda_c$.}}
    \label{fig:ablation_lambda_c}
\end{figure}

\begin{table}[!t]
\caption{Ablations to different contrastive learning and object detection-based methods. mAP performance on MS-COCO dataset is reported.}
\label{tab:ablation_othercontrastive}
\centering
\begin{tabular}{l|c}
    \toprule
    \textbf{Methods} & \textbf{mAP on MS-COCO}           \\
    \midrule
    Baseline & $69.8~\textcolor{gray}{(+0.0)}$             \\
    \midrule
    $+$BYOL & $65.5~\textcolor{red}{(-4.3)}$             \\
    $+$MoCo & $61.6~\textcolor{red}{(-8.2)}$             \\
    $+$SimCLR & $64.9~\textcolor{red}{(-4.9)}$           \\
    $+$Detco & $61.2~\textcolor{red}{(-8.6)}$ \\ 
    $+$SoCo & $58.2~\textcolor{red}{(-11.6)}$ \\
    $+$UP-DETR & $69.2~\textcolor{red}{(-0.6)}$           \\
    \midrule
    \textbf{Scob} (Ours) & $\mathbf{74.8}~\textcolor[RGB]{96,177,87}{(+5.0)}$  \\
    \bottomrule
\end{tabular}
\end{table}

\begin{figure*}[!t]
\centering
\begin{minipage}[b]{0.45\textwidth}
\centering
\includegraphics[width=0.85\linewidth]{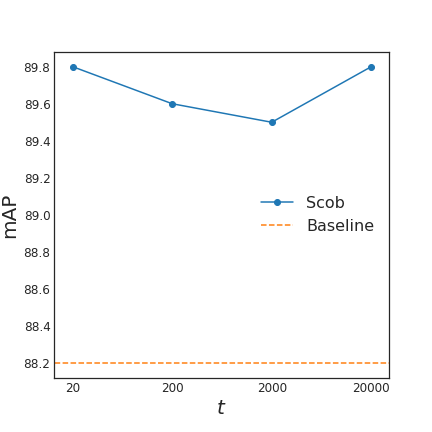}
\caption{\textbf{mAP on VOC12 of our method \textbf{Scob} and \textbf{Baseline} with different $t$ values.}}
\label{fig:ablation_k_voc}
\end{minipage}\quad\quad
\begin{minipage}[b]{0.45\textwidth}
\centering
\includegraphics[width=0.85\linewidth]{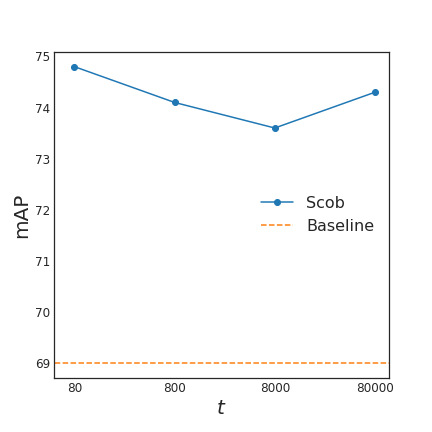}
\caption{\textbf{mAP on MS-COCO dataset of our method \textbf{Scob} and \textbf{Baseline} with different $t$ values.}}
\label{fig:ablation_k_coco}
\end{minipage}
\end{figure*}

\begin{table*}[!t]
\caption{ Ablations to semantic masks on VOC 2012 and MS-COCO dataset.  }
\label{tab:ablation_mask}
\centering
\begin{tabular}{ll|c|cccccc}
    \toprule
\textbf{Datasets}& \textbf{Methods} & \textbf{mAP}& \textbf{OP}     & \textbf{OR}     & \textbf{OF1} & \textbf{CP} &\textbf{CR}&\textbf{CF1}\\
    \midrule
    \multirow{3}{*}{VOC 2012} & Baseline & $88.2~(+0.0)$ &85.2 & 83.3 &84.2 &83.9 &81.1 &82.5      \\
     & GT Masks& $\mathbf{90.0}~\textcolor[RGB]{96,177,87}{(+1.8)}$ &86.2 & \textbf{86.2}& \textbf{86.2}& \textbf{83.9}& \textbf{84.9}& \textbf{84.4}     \\
    & CAMs& $89.8~\textcolor[RGB]{96,177,87}{(+1.6)}$ &\textbf{88.2}& 82.1& 85.0& 84.6& 81.7& 83.1 \\
\midrule
        \multirow{3}{*}{MS COCO} & Baseline & $69.8~(+0.0)$  & 70.1 & 68.3 & 69.2 & 68.0 & 63.1 & 65.5       \\
     & GT Masks& $\mathbf{74.8}~\textcolor[RGB]{96,177,87}{(+5.0)}$ &79.0 & \textbf{68.4}& \textbf{73.3}& \textbf{78.8}& \textbf{63.2}& \textbf{70.1}     \\
    & CAMs& $\mathbf{74.8}~\textcolor[RGB]{96,177,87}{(+5.0)}$ & \textbf{81.1}& 66.2 &72.9 &78.7 &61.1 &68.8  \\
    \bottomrule
\end{tabular}
\end{table*}

\textbf{Ablations to other contrastive learning and detection-based methods}  We replace our Scob with prevailing contrastive learning, BYOL~\citep{byol}, MoCo~\citep{He_2020_CVPR}, SimCLR~\citep{chen2020simple}, Detco~\citep{xie2021detco}, SoCo~\citep{xie2021unsupervised}, and object detection-based methods UP-DETR~\citep{Dai_2021_CVPR} in \tabref{tab:ablation_othercontrastive}. The baseline is taken from the w/o all setting in \tabref{tab:ablation}, which is trained without contrastive learning. As shown in the \tabref{tab:ablation_othercontrastive}, other contrastive learning methods lead to sharp performance drops compared to our approach. 
To compare with Detco~\citep{xie2021detco}, SoCo~\citep{xie2021unsupervised}, we adopt their official pre-trained model of the MS-COCO dataset and fine-tune these models following the same baseline,~\ie, ROLE~\citep{Cole_2021_CVPR}. Although these methods provide a satisfactory representation of object detection but rely on different \textbf{contrastive manners} compared to ours.
We focus on the contrastive learning of \textbf{semantic views} with CAMs to preserve the extraction of holistic objects. While Detco and SoCo focus on the relationship between \textbf{local patches or cropped local views} without the semantic constraints. The different focuses make the learned representations suitable for different tasks, \ie, recognition and detection.
The other reason is that our Scob constructed an instance priority tree for adaptively selecting negative CAMs, while Detco and SoCo follow a common contrastive trend with negative image-level representations or simply omit this relationship.

Considering that multi-label images usually contain multiple semantic objects, data augmentations used in them constantly include random cropping to create multiple views of the original images, which may unexpectedly focus on different objects in multi-label learning, introducing wrong positive or negative samples. The wrong samples make the contrastive learning fail and mislead models to distinguish objects by the wrong features, which is the main reason for the performance drops. In our scheme, we resort to object-level representation for building contrastive learning. It selects semantic objects from images with CAM guidance instead of direct data augmentation, which tackles the problem of semantic inconsistency in multi-label learning and improves the generalization ability of network models.

\textbf{Effect of different ratios $\gamma_{\textrm{cam}}$ of semantic mask} As important guidance to semantic learning, we validate different ratios $\gamma_{\textrm{cam}}$ of Eqn.~\eqref{eq:cam} in~\figref{fig:ablation_gamma_cam}. This parameter determines how responsive the mask guidance is to the activation maps of objects. A larger value can make the mask guidance filter more unrelated to noisy context, while also requiring the masked objects to be more typical. With the increase of the threshold, the mask guidance has the potential to focus on the main object better, while the thresholds larger than $0.5$ would lead to incomplete object understanding during training, especially when the model is not so confident about the recognized objects in the early steps.

\begin{table*}[!t]
\caption{ Comparisons of different optimization methods on MS-COCO dataset. Higher results are viewed in bold.}  
\label{tab:optimization2}
\setlength{\tabcolsep}{3.0mm}
\centering
\begin{tabular}{l|c|cccccc}
    \toprule
\textbf{Methods} & \textbf{mAP}& \textbf{OP}     & \textbf{OR}     & \textbf{OF1} & \textbf{CP} &\textbf{CR}&\textbf{CF1}\\
    \midrule
    Baseline & $69.8~(+0.0)$  & 70.1 & 68.3 & 69.2 & 68.0 & 63.1 & 65.5     \\
    \midrule
    Joint Optim.& $69.7~\textcolor{red}{(-0.1)}$& 75.7 & 65.8 & 70.4 & 72.8 & 59.7 & 65.6        \\
    Two Stage& $74.2~\textcolor[RGB]{96,177,87}{(+4.3)}$ &75.4 & \textbf{70.9} & \textbf{73.1} & 75.4 & \textbf{64.9} & \textbf{69.8}      \\
\textbf{Scob} (Ours) & $\mathbf{74.8}~\textcolor[RGB]{96,177,87}{(+5.0)}$ & \textbf{81.1}& 66.2 &72.9 &\textbf{78.7} &61.1 &68.8 \\
    \bottomrule
\end{tabular}
\end{table*}

\textbf{Ablation to loss weight $\lambda_c$}
For the contrastive loss weight $\lambda_c$ on term $\mathcal L_{\mathrm{cont}}$, we conduct different parameter settings with different numbers of magnitude, while this hyperparameter can also be jointly defined by adjusting the learning rate in the contrastive learning phase. As in~\figref{fig:ablation_lambda_c}, when the contrastive learning weight $\lambda_c$ increases to a proper weight,~\ie, $0.1$, the performance shows a slight improvement. However, exaggerating the effects of contrastive loss would override the main objective of multi-label classification and then lead to inaccurate CAM guidance. Note that in this paper we do not need to carefully tune these hyperparameters but validate their effectiveness under diverse scenarios of multiple datasets.

\textbf{Ablation to the number of negative samples in contrastive learning}
We validate the effect of the number of negative samples in contrastive learning. Here we present results in \figref{fig:ablation_k_voc} on the VOC dataset and \figref{fig:ablation_k_coco} on MS-COCO dataset, where the baseline result is exhibited in the orange dotted line. We select the Top-$1$, Top-$10$, Top-$100$, and Top-$1000$ confident instances in each class following the algorithms of IPT. The curves on different datasets show similar variation tendencies that selecting the most representative samples of each class would benefit the final performance while selecting more than 8,000 samples for the MS-COCO dataset would lead to a performance drop of over $1\%$. Counter-intuitively, when increasing the number of instances with sufficient training instances,~\ie, 80,000 samples in MS-COCO, the network has the potential to understand the object representations with a performance improvement. However, constraining the network to recognize a sufficiently large number of negative samples will result in a heavy computation burden,~\ie, 1000$\times$ sample numbers for learning.
Hence, in the trade-off between computational resources and performances, we respectively use $20$ and $80$ negative samples on VOC and MS-COCO datasets, by selecting the most confident instances in each class.

\textbf{Semantic CAMs \textit{vs.} ground truth masks}
To understand the influence brought by semantic masks of different qualities, we modify the CAM guidance with the ground truth segmentation masks.
In order to migrate the segmentations to our tasks, for ground truth, labels are dropped and a single positive label is retained. Next, we only keep segmentation masks that contain the objects belonging to the single positive label and ignore other segmentation annotations. The segmentation masks are resized with max pooling to generate masks while fitting the shape of the features.
The experimental results of Scob with CAM and ground truth on the two benchmark datasets are listed in \tabref{tab:ablation_mask}.
It is interesting that the gap in mAP evaluations is relatively small,~\ie, less than 0.2\%, demonstrating that the CAM selected by IPT is of high quality in contrastive learning.  We also conduct detailed analyses with six other evaluation metrics in~\tabref{tab:ablation_mask}. Models with generated CAMs show a higher overall precision (OP) but with lower overall recall (OR) compared to the models with GT masks. This implies that the GT masks can activate more features of positive classes for the final classification but meanwhile introduces additional noisy features for classification.

 \begin{table}[!t]
\caption{ CAM evaluations on feature space of different variants on the VOC training set.}
\label{tab:comp-cams}
\centering
\begin{tabular}{l|ccc}
    \toprule
\textbf{Methods}& \textbf{Precision}     & \textbf{Recall}     &\textbf{F1-Score}\\
    \midrule
    Scob w/o CL & 48.1 & 63.3 & 54.7  \\
    Scob w/ CL &  \textbf{50.0} & \textbf{68.2} & \textbf{57.7}  \\
   
    \bottomrule
\end{tabular}
\end{table}

\begin{figure}[!t]
\centering
\includegraphics[width=0.85\linewidth]{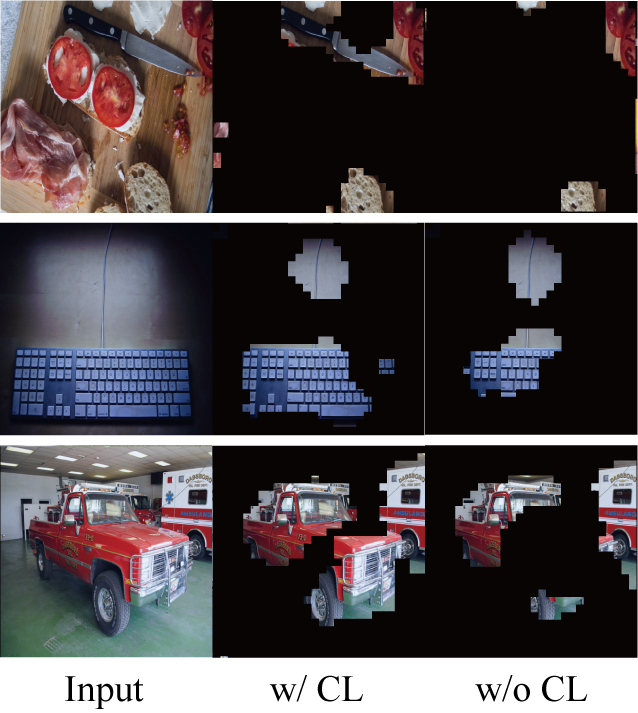}
\caption{\textbf{Visualization of semantic masks on MS-COCO.} Each group of images from left to right are \textbf{original image}, \textbf{Scob} and \textbf{Scob w/o CL} respectively.}
\label{fig:mask_loc}
\end{figure}

\textbf{Effect of EM-based Bootstrapping} To verify the effectiveness of the proposed EM-based bootstrapping, we conduct experiments with two different alternatives. 
\begin{enumerate}
\item Joint Optimization: training the network parameters $\Theta$ and object CAMs $\mathbf{M}$ simultaneously. 
\item Two Stage: training network without CAM guidance $\mathbf{M}$ in the first 10 epochs and then conducting the proposed EM-based bootstrapping in the next 20 epochs.
\end{enumerate}
In~\tabref{tab:optimization2}, the joint optimization shows similar performance with the baseline models while remaining a gap (5.1 in mAP) with the proposed EM-based bootstrapping model. This indicates the inferior initialization of CAMs and network parameters would harm the optimization process and cannot be fully corrected during learning.
Results in the third row denote this two-stage training manner, which shows a clear performance increase (+4.3\%) compared to the baseline models, but still lower than the proposed EM bootstrapping training scheme.  From~\tabref{tab:optimization2}, the method without using CAMs in the initial training phase (\ie, Two stage) would lead to a clear performance drop in overall precision (OP) but with higher recalls (OR).

\begin{table}[!t]
\caption{Comparisons of inference speed (per image) and performance on MS-COCO dataset.}  
\label{tab:speed}
\setlength{\tabcolsep}{1.0mm}
\renewcommand{\arraystretch}{1.0}
\centering
\begin{tabular}{l|ccc}
    \toprule
    \textbf{Methods} & \textbf{Speed}  & \textbf{Settings}   & \textbf{mAP}       \\
    \midrule
     Baseline~\citep{Cole_2021_CVPR}&     \textbf{$7.67ms$} & 1-positive &  69.8     \\
     SST~\citep{https://doi.org/10.48550/arxiv.2112.10941} & $19.84ms$ & 10\% partial &     68.1      \\
     SARB~\citep{Pu2022SARB} & $19.09ms$ & 10\% partial &     71.2      \\
     Scob (Ours) & $12.14ms$ & 1-positive&  \textbf{74.8} \\
    \bottomrule
\end{tabular}
\end{table}

\textbf{Time efficiency} In~\tabref{tab:speed}, we conduct experiments to compare the evaluation speed on a single NVIDIA 3090 GPU with the input resolution of $448\times 448$. The baseline model~\citep{Cole_2021_CVPR} without any additional modules shows a fast inference speed of 7.67ms per image. Compared to recent works~\ie, SST~\citep{https://doi.org/10.48550/arxiv.2112.10941}, SARB~\citep{Pu2022SARB}, our proposed method achieves a good trade-off with notable performance improvement and acceptable time costs.

\subsection{Visualization Analyses}

\textbf{Visualizations of semantic masks} In \figref{fig:mask_loc}, we visualize the instances produced by Scob (second column) and Scob w/o contrastive learning (third column). We can observe that the semantic masks generated with contrastive learning show clear silhouettes, focuses more on the semantic objects, and mitigates the influence of other contextual objects, \eg, discovering \textit{knife} out of other foods in the first row. We have also conducted a detailed comparison of CAMs with ground truth segmentation masks in the feature space, as in~\tabref{tab:comp-cams}. By adding the contrastive learning on CAMs, the F1-score has improved by 3.0\% with clearer localizations.
Object-level contrastive learning distinguishes different instances in multi-object images and improves the recognition performance.  Besides, we also exhibit more visualization results related to specific categories in~\figref{fig:more_masks} of our Appendix.

\begin{figure}[!t]
\centering
\includegraphics[width=0.85\linewidth]{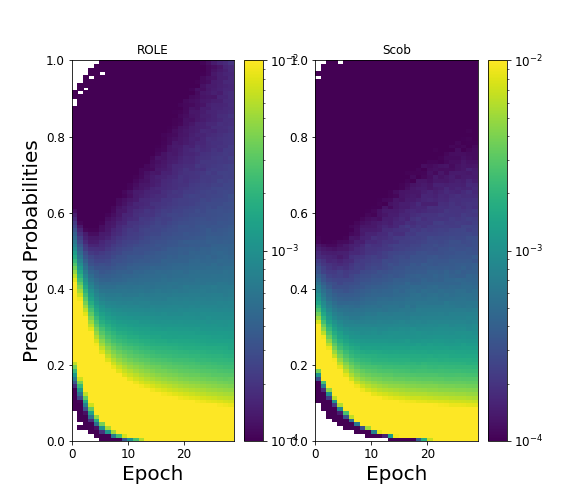}
\caption{\textbf{Distribution of predicted probabilities for unknown negatives by \textbf{ROLE} (left) and \textbf{Ours} (right) on single-positive COCO.} Each pixel stands for histograms, indicating a frequency from $1.0$ (yellow) to $0.0$ (blue) of the specific confidence value.}
\label{fig:distribution}
\end{figure}

\textbf{Distributions of unknown predictions} Here we present the distribution of predicted probabilities for unknown negatives on single-positive COCO in \figref{fig:distribution}, which are the major parts of unknown labels. During training stages, Scob fast predicts more accurate negative labels. The predicted probabilities by our approach of ground-truth negatives show a quick convergence to the $0$ side, while the probability distribution predicted by ROLE \citep{Cole_2021_CVPR} is dispersed. This indicates that due to the new module and contrastive learning, the model obtains more knowledge during training, and learns more representative features to distinguish the differences between objects so that the non-existent objects can be eliminated faster, which improves the overall performance.

\section{Conclusions and Limitations}\label{sec:conclusion}
In this paper, we focus on the problem of multi-label image recognition with single-positive incomplete annotations. To this end, we argue to use a gradient-based class activation map from the previous step as guidance and propose a semantic contrastive bootstrapping learning framework to iteratively refine the guidance and label predictions. In this framework, we propose a recurrent semantic masked transformer to extract accurate clear object-level features and a contrastive constraint with instance priority trees for building cross-object relationships. Experimental results verify our proposed method achieves superior performance compared to state-of-the-art methods.

As our proposed method is a weakly-supervised learning method, it still faces problems when discovering the full ground truth annotations, which limits the performance compared to the fully-supervised methods. Besides, on the fine-grained multi-label learning with extremely limited labels, prevailing techniques including our method remain space for further exploration, including investigating fine-grained local parts instead of the holistic objects, which we would like to investigate in our future work.

\begin{acknowledgements}
This work is partially supported by grants from the National Natural Science Foundation of China under contracts No. 62132002, No. 62202010 and 
also \\supported by China Postdoctoral Science Foundation No. 2022M710212.

\end{acknowledgements}

\section*{Appendix}

\begin{figure*}[htbp]
\centering
\includegraphics[width=0.91\linewidth]{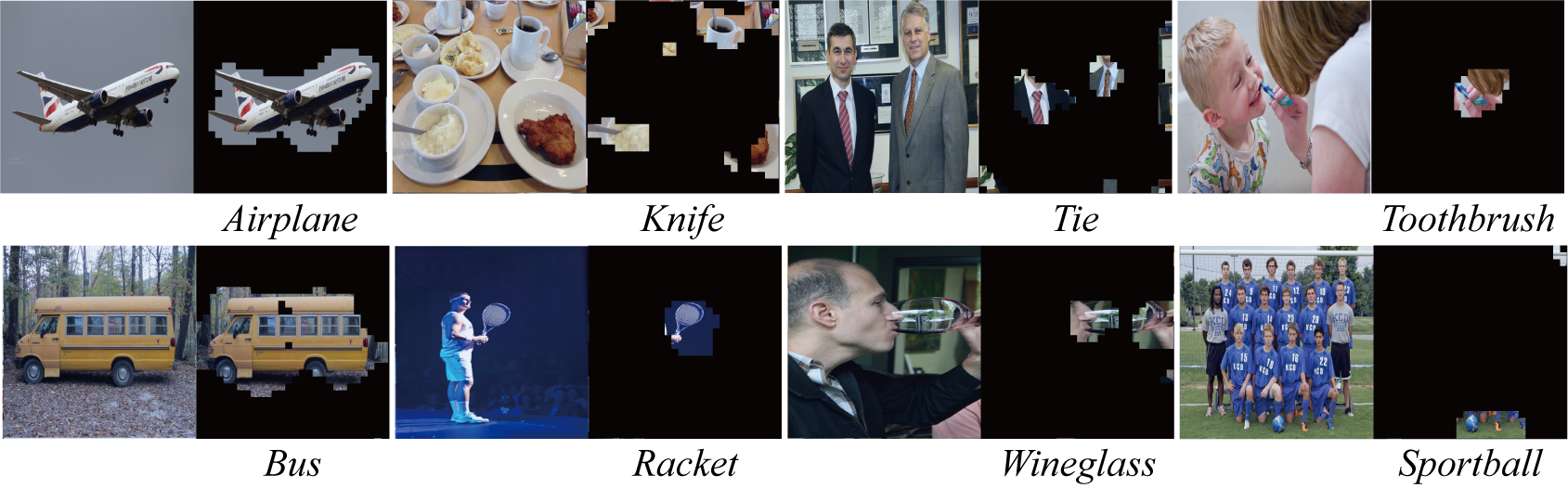}
\caption{\textbf{Visualization of semantic masks on MS-COCO.} In each group, the \textbf{left} and \textbf{right} images denote the input image and the masked class activation maps in our proposed Scob approach.}
\label{fig:more_masks}
\end{figure*}

\subsection*{A. Visualizations on Class-specific Semantic Masks}

Beyond the ablations on semantic masks in the main manuscript, here we exhibit more semantic masks generated on the MS-COCO dataset in \figref{fig:more_masks}. In each group, the left and right images denote the input image and the masked class activation maps in our proposed Scob approach. As in the first two columns, it can be found that our proposed approach focuses on the semantic objects~\eg, \textit{airplane} and \textit{bus}, while filtering the background information. In the second to the fourth groups of \figref{fig:more_masks}, two major challenges occur to distinguish these objects: 1) the semantic objects are relatively small compared to the image size; 2) these objects show high dependencies on the other objects or namely co-occurrences. As can be seen from this generated semantic guidance, our proposed Scob has the ability to distinguish the specific object from its related objects and then forms the representative features.

\begin{figure*}[t]
    \centering
    \includegraphics[width=0.9\linewidth]{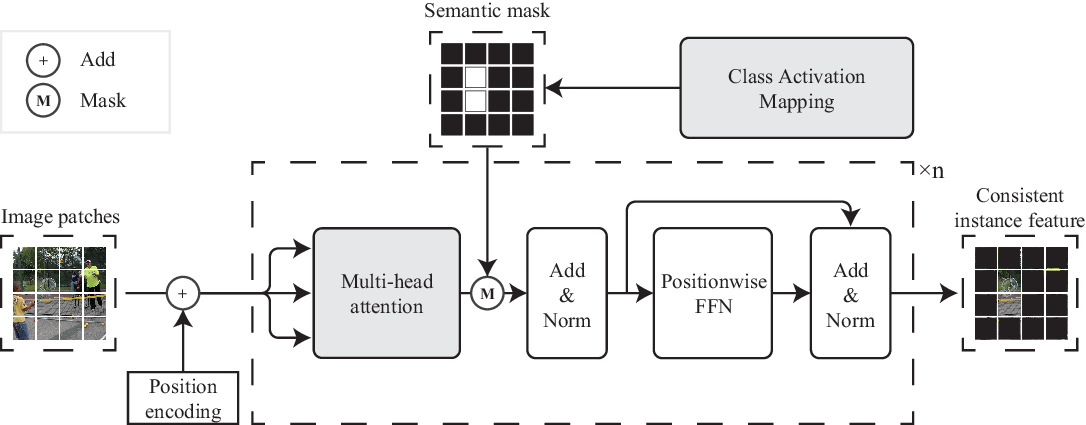}
    \caption{\textbf{The network architecture of the proposed Recurrent Semantic Masked Transformer.} We leverage the masked multi-head attention mechanisms with iterative semantic guidance generated by CAMs.  }
    \label{fig:smt}
\end{figure*}

\subsection*{B. Performances on Recognizing Small Objects}

Besides \figref{fig:more_masks} showing some CAM results of large and small-scale objects. We additionally list the AP results of 10 small-scale classes in the MS-COCO dataset to show our approach is robust to them in \tabref{tab:ap_of_10smallscale}.

\begin{table}[t]
\caption{The AP results of 10 small-scale classes in MS-COCO dataset.}
\label{tab:ap_of_10smallscale}
\centering
\begin{tabular}{ll|c}
    \toprule
    \textbf{\#} & \textbf{Classes} & \textbf{AP}           \\
    \midrule
    1 & banana & $93.0$             \\
    2 & baseball bat & $76.0$             \\
    3 & baseball glove & $80.9$             \\
    4 & book & $70.8$             \\
    5 & mouse & $74.6$             \\
    6 & remote & $84.3$             \\
    7 & scissors & $86.1$             \\
    8 & stop sign & $84.6$             \\
    9 & tie & $88.4$             \\
    10 & toothbrush & $82.4$             \\
    \bottomrule
\end{tabular}
\end{table}

\subsection*{C. Details on Recurrent Semantic Masked Transformer}\label{sec:smt}

The proposed recurrent Semantic Masked Transformer (SMT) mainly consists of positional encoding, semantic mask, and multiple multi-head attention units as in previous work \citep{NIPS2017_3f5ee243}. Different from the implementation of these works, we rely on the feature maps of ResNet backbones as image feature patches (illustrated as image patches for a better view). Inspired by the masked coding manner \citep{https://doi.org/10.48550/arxiv.1810.04805} in the field of natural language processing, here we adopt the class activation generated from the last optimization as the guidance for semantic masked attention. We present the detailed network architectures in \figref{fig:smt}. 

\textbf{Semantic masked encoding} Denote $\mathbf F=\Phi(\mathbf x;\theta_b) \in \mathbb R^{HW \times K}$ as the backbone feature extracted from ResNet stages with parameter $\theta_b \in \Theta$. $\mathbf F$ is split into $H \times W$ patches $\left\{ \mathbf F_{i,j} \mid i \in \{1,2,\dots,H\}, j \in \{1,2,\dots,W\} \right\}$ of which each patch has channel size $K$, as the input of SMT. As mentioned in Section 3.2, the recurrent SMT applies a learnable positional encoding $\Delta(\cdot):\mathbb N^{W \times H} \mapsto \mathbb R^1$ and semantic mask to $\{ \mathbf F_{i,j} \}$:
\begin{equation}
    \mathcal M(\mathbf F_{i,j})=(\mathbf F_{i,j}+\Delta(i,j))\cdot(1-\mathbf M^c),
\end{equation}
where the symbols are described as aforementioned. The implementation of $\Delta$ follows the previous Transformer architecture,~\ie, DETR \citep{detr}. Let $\Delta_h:\mathbb N^W \mapsto \mathbb R^1$ and $\Delta_v:\mathbb N^H \mapsto \mathbb R^1$ be the horizontal and vertical encoding. Then $\Delta$ is the concatenation of them:
\begin{equation}
    \Delta(i,j) = \Delta_h(i) \oplus \Delta_v(j),
\end{equation}
where $\oplus$ denotes the feature concatenation operation.
Following \cite{https://doi.org/10.48550/arxiv.1810.04805}, positional encoding and masks are applied to the query and key of multi-head attention, providing global position information and constraining the extracted features $\mathbf H_{i,j}^c$ showing high response to only a specific semantic class related to the masks.

\textbf{Multi-scale SMTs}  As the different network stages during training are sensitive to semantic objects of different scales, hence incorporating multi-scale and multi-level features can be beneficial for final feature representations, which is also one of the challenging problems in multi-label classification tasks.   
In our implementation, we propose to use two SMTs connecting to \textit{Stage 3} and \textit{Stage 4} of ResNet-50, and combine their outputs to extract image features at multiple scales. In this manner, objects of small scales are more easily to be presented in earlier network stages without a significant loss in resolutions.

\begin{algorithm}[t]
  \caption{Insert activated node $v$ to $\mathbb G$ and adjust priority tree}\label{alg:ipt_insert}
  \KwData{IPT $\mathbb G^i=(\mathcal E^i,\mathcal N^i), i \in \{1,2,3,\dots,L\}$, new node $v$ associated with label $z$.}
  \Comment{Insert activated node $v$.}
  $\mathcal N^z \gets \mathcal N^z \cup \{v\}$\;
  \Comment{Adjust priority tree.}
  $i \gets |\mathcal N^z|-1$\;
  Choose elements $[u_0,u_1,u_2,\dots,u_n]$ in $\mathcal N^z$\;
  \While{$i>0 ~\textbf{and}~s_{\lfloor i/2 \rfloor} < s_{i}$}{
    Swap $u_{\lfloor i/2 \rfloor}$ and $u_i$\;
    $i \gets \lfloor i/2 \rfloor$\;
  }
  Adjust $\mathcal N^z$ by removing exceed nodes from the end of it\;
\end{algorithm}

\begin{algorithm}[t]
  \caption{Pop top nodes $\mathcal N'$ from $\mathbb G^i$ for negative sampling}\label{alg:ipt_pop}
  \KwData{IPT $\mathbb G^i=(\mathcal E^i,\mathcal N^i), i \in \{1,2,3,\dots,L\}$, positive label $z$, number of negative samples $t$.}
  \KwResult{Negative samples $\mathcal N'$.}
  \If{$i = z$}{
    $\mathcal N' \gets \{\}$\;
    return\;
  }
  Choose elements $[u_0,u_1,u_2,\dots,u_n]$ in $\mathcal N^i$\;
  \eIf{$|\mathcal N^z| <= t$}{
    \Comment{If the size of $\mathcal N^z$ is less than $t$, simply construct samples with all nodes.}
    $\mathcal N' \gets \mathcal N^z$\;
  }{
    \Comment{Pop top elements $t$ times to construct samples.}
    $\mathcal N' \gets \{\}$\;
    
    \For{$i\leftarrow 1$ \KwTo $t$}{
      $j \gets 0$\;
      Swap $u_j$ and $u_n$\;
      \While{$2j<|\mathcal N^z|$}{
        \eIf{$s_{2j}>s_{2j+1}$}{
          Swap $u_j$ and $u_{2j}$\;
          $j \gets 2j$\;
        }{
          Swap $u_j$ and $u_{2j+1}$\;
          $j \gets 2j+1$\;
        }
      }
      $\mathcal N' \gets \mathcal N' \cup \{u_j\}$\;
      $\mathcal N^z \gets \mathcal N^z \backslash \{u_n\}$\;
    }
    \Comment{Restore the original $\mathcal N^z$ for future operations.}
    Insert $\mathcal N'$ into $\mathcal N^z$\;
  }
\end{algorithm}

\subsection*{E. Algorithms of Instance Priority Trees}\label{sec:ipt}

Instance Priority Tree (IPT) $\mathbb{G}=\{\mathcal{E}, \mathcal{N}\}$ is a heap implemented with a complete binary tree structure, maintaining high confident instances to construct negative samples for contrastive learning. Each element $u \in \mathcal N$ is a binary tuple $(\widetilde{H}_u, s_u)$, where $\widetilde{H}_u$ is a semantic masked feature and $s_u$ is the corresponding activation confidence score. Each edge $e \in \mathcal{E}$ indicates an affiliation relationship $(u,v)$ where the parent node $u$ is with higher confidence than the leaf ones $v$.

In our implementations, we adopt an array of length $n$, $[u_0,u_1,u_2,\dots,u_n]$, to store the nodes $\mathcal N$ and describe the edges $\mathcal E$ by their indices in the array,~\ie, $u_i$ is the parent node of $u_{2i}$ and $u_{2i+1}$. The success of IPT relies on three typical operations: ``Insert activated nodes'', ``Adjust priority tree'', and ``Pop top nodes''.
In our implementation, the tree adjustment operation needs to be conducted after every \textit{Insert} or \textit{Pop} operation for maintaining the tree structures. In other words, the ``Insert activated nodes'' and ``Adjust priority tree'' functions are operated together. Here we elaborate the detailed algorithms in \algref{alg:ipt_insert} and \algref{alg:ipt_pop}.

In addition, we maintain the size of instance priority tree $\mathbb{G}$ in a proper range to reduce computation costs. In this manner, instances with lower confidence features would not be added to our tree storage, indicating the efficiency of our model.

%

\bibliographystyle{spbasic}      
\bibliography{scob}   

\begin{thebibliography}{68}
\providecommand{\natexlab}[1]{#1}
\providecommand{\url}[1]{{#1}}
\providecommand{\urlprefix}{URL }
\expandafter\ifx\csname urlstyle\endcsname\relax
  \providecommand{\doi}[1]{DOI~\discretionary{}{}{}#1}\else
  \providecommand{\doi}{DOI~\discretionary{}{}{}\begingroup
  \urlstyle{rm}\Url}\fi
\providecommand{\eprint}[2][]{\url{#2}}

\bibitem[{Balcan and Sharma(2021)}]{NEURIPS2021_7c93ebe8}
Balcan MFF, Sharma D (2021) Data driven semi-supervised learning. In: Advances
  in Neural Information Processing Systems (NeurIPS)

\bibitem[{Bar et~al.(2022)Bar, Wang, Kantorov, Reed, Herzig, Chechik, Rohrbach,
  Darrell, and Globerson}]{bar2022detreg}
Bar A, Wang X, Kantorov V, Reed CJ, Herzig R, Chechik G, Rohrbach A, Darrell T,
  Globerson A (2022) Detreg: Unsupervised pretraining with region priors for
  object detection. In: Proceedings of the IEEE/CVF Conference on Computer
  Vision and Pattern Recognition, pp 14605--14615

\bibitem[{Bucak et~al.(2011)Bucak, Jin, and Jain}]{5995734}
Bucak SS, Jin R, Jain AK (2011) Multi-label learning with incomplete class
  assignments. In: 2011 IEEE Conference on Computer Vision and Pattern
  Recognition (CVPR), \doi{10.1109/CVPR.2011.5995734}

\bibitem[{Cabral et~al.(2011)Cabral, Torre, Costeira, and
  Bernardino}]{NIPS2011_65a99bb7}
Cabral R, Torre F, Costeira JP, Bernardino A (2011) Matrix completion for
  multi-label image classification. In: Advances in Neural Information
  Processing Systems (NeurIPS)

\bibitem[{Carion et~al.(2020)Carion, Massa, Synnaeve, Usunier, Kirillov, and
  Zagoruyko}]{detr}
Carion N, Massa F, Synnaeve G, Usunier N, Kirillov A, Zagoruyko S (2020)
  End-to-end object detection with transformers. In: European Conference on
  Computer Vision (ECCV), \doi{10.1007/978-3-030-58452-8\_13}

\bibitem[{Chattopadhay et~al.(2018)Chattopadhay, Sarkar, Howlader, and
  Balasubramanian}]{gradcamplus}
Chattopadhay A, Sarkar A, Howlader P, Balasubramanian VN (2018) Grad-cam++:
  Generalized gradient-based visual explanations for deep convolutional
  networks. In: 2018 IEEE Winter Conference on Applications of Computer Vision
  (WACV), \doi{10.1109/WACV.2018.00097}

\bibitem[{Chen et~al.(2019{\natexlab{a}})Chen, Xu, Hui, Wu, and
  Lin}]{Chen_2019_ICCV}
Chen T, Xu M, Hui X, Wu H, Lin L (2019{\natexlab{a}}) Learning
  semantic-specific graph representation for multi-label image recognition. In:
  Proceedings of the IEEE/CVF International Conference on Computer Vision
  (ICCV)

\bibitem[{Chen et~al.(2020)Chen, Kornblith, Norouzi, and
  Hinton}]{chen2020simple}
Chen T, Kornblith S, Norouzi M, Hinton G (2020) A simple framework for
  contrastive learning of visual representations. In: Proceedings of the 37th
  International Conference on Machine Learning (ICML)

\bibitem[{Chen et~al.(2021)Chen, Pu, Wu, Xie, and
  Lin}]{https://doi.org/10.48550/arxiv.2112.10941}
Chen T, Pu T, Wu H, Xie Y, Lin L (2021) Structured semantic transfer for
  multi-label recognition with partial labels. arXiv preprint arXiv:211210941
  \doi{10.48550/ARXIV.2112.10941}

\bibitem[{Chen et~al.(2022)Chen, Lin, Chen, Hui, and Wu}]{9207855}
Chen T, Lin L, Chen R, Hui X, Wu H (2022) Knowledge-guided multi-label few-shot
  learning for general image recognition. IEEE Transactions on Pattern Analysis
  \& Machine Intelligence

\bibitem[{Chen et~al.(2019{\natexlab{b}})Chen, Wei, Wang, and
  Guo}]{Chen_2019_CVPR}
Chen ZM, Wei XS, Wang P, Guo Y (2019{\natexlab{b}}) Multi-label image
  recognition with graph convolutional networks. In: Proceedings of the
  IEEE/CVF Conference on Computer Vision and Pattern Recognition (CVPR)

\bibitem[{Chu et~al.(2021)Chu, Tian, Wang, Zhang, Ren, Wei, Xia, and
  Shen}]{NEURIPS2021_4e0928de}
Chu X, Tian Z, Wang Y, Zhang B, Ren H, Wei X, Xia H, Shen C (2021) Twins:
  Revisiting the design of spatial attention in vision transformers. In:
  Advances in Neural Information Processing Systems (NeurIPS)

\bibitem[{Cole et~al.(2021)Cole, Mac~Aodha, Lorieul, Perona, Morris, and
  Jojic}]{Cole_2021_CVPR}
Cole E, Mac~Aodha O, Lorieul T, Perona P, Morris D, Jojic N (2021) Multi-label
  learning from single positive labels. In: Proceedings of the IEEE/CVF
  Conference on Computer Vision and Pattern Recognition (CVPR)

\bibitem[{Dai et~al.(2021)Dai, Cai, Lin, and Chen}]{Dai_2021_CVPR}
Dai Z, Cai B, Lin Y, Chen J (2021) Up-detr: Unsupervised pre-training for
  object detection with transformers. In: Proceedings of the IEEE/CVF
  Conference on Computer Vision and Pattern Recognition (CVPR), pp 1601--1610

\bibitem[{Deng et~al.(2009)Deng, Dong, Socher, Li, Li, and
  Fei-Fei}]{deng2009imagenet}
Deng J, Dong W, Socher R, Li LJ, Li K, Fei-Fei L (2009) Imagenet: A large-scale
  hierarchical image database. In: Proceedings of the IEEE/CVF Conference on
  Computer Vision and Pattern Recognition (CVPR)

\bibitem[{Devlin et~al.(2019)Devlin, Chang, Lee, and
  Toutanova}]{https://doi.org/10.48550/arxiv.1810.04805}
Devlin J, Chang M, Lee K, Toutanova K (2019) {BERT:} pre-training of deep
  bidirectional transformers for language understanding. In: Proceedings of the
  2019 Conference of the North American Chapter of the Association for
  Computational Linguistics: Human Language Technologies

\bibitem[{Dosovitskiy et~al.(2021{\natexlab{a}})Dosovitskiy, Beyer, Kolesnikov,
  Weissenborn, Zhai, Unterthiner, Dehghani, Minderer, Heigold, Gelly,
  Uszkoreit, and Houlsby}]{dosovitskiy2021an}
Dosovitskiy A, Beyer L, Kolesnikov A, Weissenborn D, Zhai X, Unterthiner T,
  Dehghani M, Minderer M, Heigold G, Gelly S, Uszkoreit J, Houlsby N
  (2021{\natexlab{a}}) An image is worth 16x16 words: Transformers for image
  recognition at scale. In: International Conference on Learning
  Representations

\bibitem[{Dosovitskiy et~al.(2021{\natexlab{b}})Dosovitskiy, Beyer, Kolesnikov,
  Weissenborn, Zhai, Unterthiner, Dehghani, Minderer, Heigold, Gelly,
  Uszkoreit, and Houlsby}]{ViT}
Dosovitskiy A, Beyer L, Kolesnikov A, Weissenborn D, Zhai X, Unterthiner T,
  Dehghani M, Minderer M, Heigold G, Gelly S, Uszkoreit J, Houlsby N
  (2021{\natexlab{b}}) An image is worth 16x16 words: Transformers for image
  recognition at scale. In: International Conference on Learning
  Representations

\bibitem[{Durand et~al.(2019)Durand, Mehrasa, and Mori}]{Durand_2019_CVPR}
Durand T, Mehrasa N, Mori G (2019) Learning a deep convnet for multi-label
  classification with partial labels. In: Proceedings of the IEEE/CVF
  Conference on Computer Vision and Pattern Recognition (CVPR)

\bibitem[{Everingham et~al.(2007)Everingham, Van~Gool, Williams, Winn, and
  Zisserman}]{pascal-voc-2007}
Everingham M, Van~Gool L, Williams CKI, Winn J, Zisserman A (2007) The {PASCAL}
  {V}isual {O}bject {C}lasses {C}hallenge 2007 {(VOC2007)} {R}esults.
  http://www.pascal-network.org/challenges/VOC/voc2007/workshop/index.html

\bibitem[{Everingham et~al.(2012)Everingham, Van~Gool, Williams, Winn, and
  Zisserman}]{pascal-voc-2012}
Everingham M, Van~Gool L, Williams CKI, Winn J, Zisserman A (2012) The {PASCAL}
  {V}isual {O}bject {C}lasses {C}hallenge 2012 {(VOC2012)} {R}esults.
  http://www.pascal-network.org/challenges/VOC/voc2012/workshop/index.html

\bibitem[{Gao and Zhou(2021)}]{Gao2021LearningTD}
Gao BB, Zhou HY (2021) Learning to discover multi-class attentional regions for
  multi-label image recognition. IEEE Transactions on Image Processing
  30:5920--5932

\bibitem[{Ge et~al.(2018)Ge, Yang, and Yu}]{Ge_2018_CVPR}
Ge W, Yang S, Yu Y (2018) Multi-evidence filtering and fusion for multi-label
  classification, object detection and semantic segmentation based on weakly
  supervised learning. In: Proceedings of the IEEE Conference on Computer
  Vision and Pattern Recognition (CVPR)

\bibitem[{Gong et~al.(2021)Gong, Yuan, and Bao}]{NEURIPS2021_217c0e01}
Gong X, Yuan D, Bao W (2021) Understanding partial multi-label learning via
  mutual information. In: Advances in Neural Information Processing Systems
  (NeurIPS)

\bibitem[{Grill et~al.(2020)Grill, Strub, Altch\'{e}, Tallec, Richemond,
  Buchatskaya, Doersch, Avila~Pires, Guo, Gheshlaghi~Azar, Piot, kavukcuoglu,
  Munos, and Valko}]{byol}
Grill JB, Strub F, Altch\'{e} F, Tallec C, Richemond P, Buchatskaya E, Doersch
  C, Avila~Pires B, Guo Z, Gheshlaghi~Azar M, Piot B, kavukcuoglu k, Munos R,
  Valko M (2020) Bootstrap your own latent - a new approach to self-supervised
  learning. In: Advances in Neural Information Processing Systems (NeurIPS)

\bibitem[{Guo and Wang(2021)}]{Guo_2021_CVPR}
Guo H, Wang S (2021) Long-tailed multi-label visual recognition by
  collaborative training on uniform and re-balanced samplings. In: Proceedings
  of the IEEE/CVF Conference on Computer Vision and Pattern Recognition (CVPR)

\bibitem[{He et~al.(2016)He, Zhang, Ren, and Sun}]{He_2016_CVPR}
He K, Zhang X, Ren S, Sun J (2016) Deep residual learning for image
  recognition. In: Proceedings of the IEEE Conference on Computer Vision and
  Pattern Recognition (CVPR)

\bibitem[{He et~al.(2020)He, Fan, Wu, Xie, and Girshick}]{He_2020_CVPR}
He K, Fan H, Wu Y, Xie S, Girshick R (2020) Momentum contrast for unsupervised
  visual representation learning. In: IEEE/CVF Conference on Computer Vision
  and Pattern Recognition (CVPR)

\bibitem[{He et~al.(2021)He, Chen, Xie, Li, Doll{\'a}r, and
  Girshick}]{he2021masked}
He K, Chen X, Xie S, Li Y, Doll{\'a}r P, Girshick R (2021) Masked autoencoders
  are scalable vision learners. arXiv preprint arXiv:211106377

\bibitem[{Hu et~al.(2020)Hu, Xie, Du, Hong, and Tian}]{NEURIPS2020_05f971b5}
Hu H, Xie L, Du Z, Hong R, Tian Q (2020) One-bit supervision for image
  classification. In: Advances in Neural Information Processing Systems
  (NeurIPS)

\bibitem[{Huynh and Elhamifar(2020)}]{Huynh_2020_CVPR}
Huynh D, Elhamifar E (2020) Interactive multi-label cnn learning with partial
  labels. In: Proceedings of the IEEE/CVF Conference on Computer Vision and
  Pattern Recognition (CVPR)

\bibitem[{Jia et~al.(2021)Jia, Chen, and Huang}]{Jia_2021_ICCV}
Jia J, Chen X, Huang K (2021) Spatial and semantic consistency regularizations
  for pedestrian attribute recognition. In: Proceedings of the IEEE/CVF
  International Conference on Computer Vision (ICCV)

\bibitem[{Jiang et~al.(2018)Jiang, Zhou, Leung, Li, and
  Fei-Fei}]{JIANG_2018_ICML}
Jiang L, Zhou Z, Leung T, Li LJ, Fei-Fei L (2018) {M}entor{N}et: Learning
  data-driven curriculum for very deep neural networks on corrupted labels. In:
  Proceedings of the 35th International Conference on Machine Learning (ICML)

\bibitem[{Khosla et~al.(2020)Khosla, Teterwak, Wang, Sarna, Tian, Isola,
  Maschinot, Liu, and Krishnan}]{NEURIPS2020_d89a66c7}
Khosla P, Teterwak P, Wang C, Sarna A, Tian Y, Isola P, Maschinot A, Liu C,
  Krishnan D (2020) Supervised contrastive learning. In: Advances in Neural
  Information Processing Systems (NeurIPS)

\bibitem[{Li et~al.(2021)Li, Chen, Yang, Li, Zhu, Zhao, Deng, Wu, Zhao, Tang,
  and Wang}]{NEURIPS2021_6dbbe6ab}
Li Z, Chen Z, Yang F, Li W, Zhu Y, Zhao C, Deng R, Wu L, Zhao R, Tang M, Wang J
  (2021) Mst: Masked self-supervised transformer for visual representation. In:
  Advances in Neural Information Processing Systems (NeurIPS)

\bibitem[{Li et~al.(2022)Li, Zhu, Yang, Li, Zhao, Chen, Chen, Xie, Wu, Zhao,
  Tang, and Wang}]{https://doi.org/10.48550/arxiv.2203.06965}
Li Z, Zhu Y, Yang F, Li W, Zhao C, Chen Y, Chen Z, Xie J, Wu L, Zhao R, Tang M,
  Wang J (2022) Univip: A unified framework for self-supervised visual
  pre-training. arXiv preprint arXiv:220306965 \doi{10.48550/ARXIV.2203.06965}

\bibitem[{Lin et~al.(2014)Lin, Maire, Belongie, Hays, Perona, Ramanan,
  Doll{\'a}r, and Zitnick}]{10.1007/978-3-319-10602-1_48}
Lin TY, Maire M, Belongie S, Hays J, Perona P, Ramanan D, Doll{\'a}r P, Zitnick
  CL (2014) Microsoft coco: Common objects in context. In: European Conference
  on Computer Vision (ECCV)

\bibitem[{Liu et~al.(2019)Liu, Wang, Shan, and Chen}]{Liu2019}
Liu H, Wang R, Shan S, Chen X (2019) Deep supervised hashing for fast image
  retrieval. International Journal of Computer Vision 127(9):1217--1234,
  \doi{10.1007/s11263-019-01174-4}

\bibitem[{Liu et~al.(2018)Liu, Sheng, Shao, Yan, Xiang, and
  Pan}]{10.1145/3240508.3240567}
Liu Y, Sheng L, Shao J, Yan J, Xiang S, Pan C (2018) Multi-label image
  classification via knowledge distillation from weakly-supervised detection.
  In: Proceedings of the 26th ACM International Conference on Multimedia

\bibitem[{Liu et~al.(2021)Liu, Lin, Cao, Hu, Wei, Zhang, Lin, and
  Guo}]{https://doi.org/10.48550/arxiv.2103.14030}
Liu Z, Lin Y, Cao Y, Hu H, Wei Y, Zhang Z, Lin S, Guo B (2021) Swin
  transformer: Hierarchical vision transformer using shifted windows. In:
  Proceedings of the IEEE/CVF International Conference on Computer Vision
  (ICCV)

\bibitem[{Oord et~al.(2018)Oord, Li, and Vinyals}]{Oord2018RepresentationLW}
Oord Avd, Li Y, Vinyals O (2018) Representation learning with contrastive
  predictive coding. arXiv preprint arXiv:180703748

\bibitem[{Pu et~al.(2022)Pu, Chen, Wu, and Lin}]{Pu2022SARB}
Pu T, Chen T, Wu H, Lin L (2022) Semantic-aware representation blending for
  multi-label image recognition with partial labels. In: Proceedings of the
  AAAI Conference on Artificial Intelligence

\bibitem[{Rao et~al.(2021)Rao, Zhao, Zhu, Lu, and Zhou}]{NEURIPS2021_07e87c2f}
Rao Y, Zhao W, Zhu Z, Lu J, Zhou J (2021) Global filter networks for image
  classification. In: Advances in Neural Information Processing Systems
  (NeurIPS)

\bibitem[{Sener and Koltun(2018)}]{NEURIPS2018_432aca3a}
Sener O, Koltun V (2018) Multi-task learning as multi-objective optimization.
  In: Advances in Neural Information Processing Systems (NeurIPS)

\bibitem[{Shao et~al.(2021)Shao, Bian, Chen, Wang, Zhang, Ji, and
  zhang}]{NEURIPS2021_10c272d0}
Shao Z, Bian H, Chen Y, Wang Y, Zhang J, Ji X, zhang y (2021) Transmil:
  Transformer based correlated multiple instance learning for whole slide image
  classification. In: Advances in Neural Information Processing Systems
  (NeurIPS)

\bibitem[{Shin et~al.(2022)Shin, Ishii, and Narihira}]{Shin2022}
Shin A, Ishii M, Narihira T (2022) Perspectives and prospects on transformer
  architecture for cross-modal tasks with language and vision. International
  Journal of Computer Vision 130(2):435--454, \doi{10.1007/s11263-021-01547-8}

\bibitem[{Song et~al.(2021)Song, Liu, Sun, and Shang}]{Song2021}
Song L, Liu J, Sun M, Shang X (2021) Weakly supervised group mask network for
  object detection. International Journal of Computer Vision 129(3):681--702,
  \doi{10.1007/s11263-020-01397-w}

\bibitem[{Tsipras et~al.(2020)Tsipras, Santurkar, Engstrom, Ilyas, and
  Madry}]{Dimitris_2020_ICML}
Tsipras D, Santurkar S, Engstrom L, Ilyas A, Madry A (2020) From imagenet to
  image classification: Contextualizing progress on benchmarks. In: Proceedings
  of the 37th International Conference on Machine Learning (ICML)

\bibitem[{Tsoumakas and Katakis(2009)}]{Tsoumakas_2009_MLOverview}
Tsoumakas G, Katakis I (2009) Multi-label classification: An overview.
  International Journal of Data Warehousing and Mining
  \doi{10.4018/jdwm.2007070101}

\bibitem[{Vaswani et~al.(2017)Vaswani, Shazeer, Parmar, Uszkoreit, Jones,
  Gomez, Kaiser, and Polosukhin}]{NIPS2017_3f5ee243}
Vaswani A, Shazeer N, Parmar N, Uszkoreit J, Jones L, Gomez AN, Kaiser Lu,
  Polosukhin I (2017) Attention is all you need. In: Advances in Neural
  Information Processing Systems (NeurIPS)

\bibitem[{Wah et~al.(2011)Wah, Branson, Welinder, Perona, and
  Belongie}]{WahCUB_200_2011}
Wah C, Branson S, Welinder P, Perona P, Belongie S (2011) The caltech-ucsd
  birds-200-2011 dataset. Tech. Rep. CNS-TR-2011-001, California Institute of
  Technology

\bibitem[{Wang et~al.(2022{\natexlab{a}})Wang, Xiao, Li, Feng, Niu, Chen, and
  Zhao}]{https://doi.org/10.48550/arxiv.2201.08984}
Wang H, Xiao R, Li Y, Feng L, Niu G, Chen G, Zhao J (2022{\natexlab{a}}) Pico:
  Contrastive label disambiguation for partial label learning. arXiv preprint
  arXiv:220108984 \doi{10.48550/ARXIV.2201.08984}

\bibitem[{Wang et~al.(2016)Wang, Yang, Mao, Huang, Huang, and Xu}]{7780620}
Wang J, Yang Y, Mao J, Huang Z, Huang C, Xu W (2016) Cnn-rnn: A unified
  framework for multi-label image classification. In: 2016 IEEE Conference on
  Computer Vision and Pattern Recognition (CVPR), \doi{10.1109/CVPR.2016.251}

\bibitem[{Wang et~al.(2022{\natexlab{b}})Wang, Yu, De~Mello, Kautz, Anandkumar,
  Shen, and Alvarez}]{wang2022freesolo}
Wang X, Yu Z, De~Mello S, Kautz J, Anandkumar A, Shen C, Alvarez JM
  (2022{\natexlab{b}}) Freesolo: Learning to segment objects without
  annotations. In: Proceedings of the IEEE/CVF Conference on Computer Vision
  and Pattern Recognition, pp 14176--14186

\bibitem[{Wang et~al.(2021)Wang, Huang, Song, Huang, and
  Huang}]{NEURIPS2021_64517d84}
Wang Y, Huang R, Song S, Huang Z, Huang G (2021) Not all images are worth 16x16
  words: Dynamic transformers for efficient image recognition. In: Advances in
  Neural Information Processing Systems (NeurIPS)

\bibitem[{Wolfe et~al.(2005)Wolfe, Horowitz, and Kenner}]{Wolfe2005}
Wolfe JM, Horowitz TS, Kenner NM (2005) Rare items often missed in visual
  searches. Nature \doi{10.1038/435439a}

\bibitem[{Wu et~al.(2018)Wu, Jia, Liu, Ghanem, and Lyu}]{wu2018multi}
Wu B, Jia F, Liu W, Ghanem B, Lyu S (2018) Multi-label learning with missing
  labels using mixed dependency graphs. International Journal of Computer
  Vision

\bibitem[{Wu et~al.(2019)Wu, Chen, Fan, Zhang, Hou, Liu, and Zhang}]{Wu_2019}
Wu B, Chen W, Fan Y, Zhang Y, Hou J, Liu J, Zhang T (2019) Tencent {ML}-images:
  A large-scale multi-label image database for visual representation learning.
  {IEEE} Access \doi{10.1109/access.2019.2956775}

\bibitem[{Xie et~al.(2021{\natexlab{a}})Xie, Ding, Wang, Zhan, Xu, Sun, Li, and
  Luo}]{xie2021detco}
Xie E, Ding J, Wang W, Zhan X, Xu H, Sun P, Li Z, Luo P (2021{\natexlab{a}})
  Detco: Unsupervised contrastive learning for object detection. In:
  Proceedings of the IEEE/CVF International Conference on Computer Vision, pp
  8392--8401

\bibitem[{Xie et~al.(2021{\natexlab{b}})Xie, Zhan, Liu, Ong, and
  Loy}]{xie2021unsupervised}
Xie J, Zhan X, Liu Z, Ong YS, Loy CC (2021{\natexlab{b}}) Unsupervised
  object-level representation learning from scene images. Advances in Neural
  Information Processing Systems 34:28864--28876

\bibitem[{Xu et~al.(2013)Xu, Jin, and Zhou}]{NIPS2013_e58cc5ca}
Xu M, Jin R, Zhou ZH (2013) Speedup matrix completion with side information:
  Application to multi-label learning. In: Advances in Neural Information
  Processing Systems (NeurIPS)

\bibitem[{Yang et~al.(2016)Yang, Zhou, and Cai}]{10.1007/978-3-319-46448-0_50}
Yang H, Zhou JT, Cai J (2016) Improving multi-label learning with missing
  labels by structured semantic correlations. In: European Conference on
  Computer Vision (ECCV)

\bibitem[{Yuan et~al.(2021)Yuan, Fu, Huang, Lin, Zhang, Chen, and
  Wang}]{NEURIPS2021_3bbfdde8}
Yuan Y, Fu R, Huang L, Lin W, Zhang C, Chen X, Wang J (2021) Hrformer:
  High-resolution vision transformer for dense predict. In: Advances in Neural
  Information Processing Systems (NeurIPS)

\bibitem[{Yun et~al.(2021)Yun, Oh, Heo, Han, Choe, and Chun}]{Yun_2021_CVPR}
Yun S, Oh SJ, Heo B, Han D, Choe J, Chun S (2021) Re-labeling imagenet: From
  single to multi-labels, from global to localized labels. In: Proceedings of
  the IEEE/CVF Conference on Computer Vision and Pattern Recognition (CVPR)

\bibitem[{Zhang et~al.(2021{\natexlab{a}})Zhang, Wang, Hou, WU, Wang, Okumura,
  and Shinozaki}]{NEURIPS2021_995693c1}
Zhang B, Wang Y, Hou W, WU H, Wang J, Okumura M, Shinozaki T
  (2021{\natexlab{a}}) Flexmatch: Boosting semi-supervised learning with
  curriculum pseudo labeling. In: Advances in Neural Information Processing
  Systems (NeurIPS)

\bibitem[{Zhang et~al.(2019)Zhang, Han, Zhao, and Meng}]{Zhang2019}
Zhang D, Han J, Zhao L, Meng D (2019) Leveraging prior-knowledge for weakly
  supervised object detection under a collaborative self-paced curriculum
  learning framework. International Journal of Computer Vision 127(4):363--380,
  \doi{10.1007/s11263-018-1112-4}

\bibitem[{Zhang et~al.(2021{\natexlab{b}})Zhang, Pang, Chen, and
  Loy}]{NEURIPS2021_55a7cf9c}
Zhang W, Pang J, Chen K, Loy CC (2021{\natexlab{b}}) K-net: Towards unified
  image segmentation. In: Advances in Neural Information Processing Systems
  (NeurIPS)

\bibitem[{Zhao et~al.(2021)Zhao, Yan, Zhao, Guo, Huang, and
  Li}]{Zhao_2021_ICCV}
Zhao J, Yan K, Zhao Y, Guo X, Huang F, Li J (2021) Transformer-based dual
  relation graph for multi-label image recognition. In: Proceedings of the
  IEEE/CVF International Conference on Computer Vision (ICCV)

\end{thebibliography}

\section*{Data Availability Statement}

The datasets generated during and/or analyzed during the current study are available in the original references,~\ie, PASCAL VOC 2007/2012~\citep{pascal-voc-2012}~\url{http://host.robots.ox.ac.uk/pascal/VOC/}, Microsoft COCO 2014~\citep{10.1007/978-3-319-10602-1_48}~\url{https://cocodataset.org/}, CUB-200-2011~\citep{WahCUB_200_2011}~\url{https://www.vision.caltech.edu/datasets/cub_200_2011/}. The source codes and models corresponding to this study are publicly available. 

\balance
\end{document}